\pdfoutput=1

\documentclass[11pt, dvipsnames, table]{article}

\usepackage{acl}
\usepackage{times}
\usepackage{latexsym}
\usepackage[T1]{fontenc}
\usepackage[utf8]{inputenc}
\usepackage{microtype}
\usepackage{inconsolata}
\usepackage{amsmath}
\usepackage{amssymb}
\usepackage{amsthm}
\usepackage{graphicx}
\usepackage{mathtools}
\usepackage{listings}
\usepackage{tikz}
\usepackage{siunitx}
\usepackage{tipa}
\usepackage{float}
\usepackage{bbm}
\usepackage{breqn}
\usepackage{utils}
\usepackage{tikz-dependency}
\usepackage{nicefrac}

\usepackage{caption}
\usepackage{subcaption}
\usepackage{linguex}

\usepackage{todonotes}

\makeatletter
\newcommand*\iftodonotes{\if@todonotes@disabled\expandafter\@secondoftwo\else\expandafter\@firstoftwo\fi}  
\makeatother

\newcommand{\aann}{\textsc{aann}}
\newcommand{\aanns}{\textsc{aann}s}
\newcommand{\anan}{\textsc{anan}}
\newcommand{\anans}{\textsc{anan}s}
\newcommand{\naan}{\textsc{naan}}
\newcommand{\naans}{\textsc{naan}s}
\newcommand{\noaann}{\textsc{no aann}}
\newcommand{\slor}{\textsc{slor}}
\newcommand{\construction}{\mathcal{C}}

\definecolor{blu}{HTML}{3468C0}
\definecolor{yello}{HTML}{e6ab02}
\definecolor{piink}{HTML}{e7298a}
\definecolor{greeen}{HTML}{66a61e}
\definecolor{reed}{HTML}{b2182b}
\definecolor{firebrick}{HTML}{B22222}
\definecolor{steelblue}{HTML}{4682B4}
\definecolor{puuurpl}{HTML}{B5739D}
\definecolor{gren}{HTML}{1b9e77}
\definecolor{prpl}{HTML}{7570b3}
\definecolor{orang}{HTML}{d95f02}
\definecolor{vlightgreen}{HTML}{82B366}
\definecolor{vviolet}{HTML}{56517E}

\definecolor{regexv1}{HTML}{B46504}
\definecolor{regexv2}{HTML}{56517E}
\definecolor{regexv3}{HTML}{0E8088}

\definecolor{naanpollution}{HTML}{488795}
\definecolor{ananpollution}{HTML}{ae4d9f}

\usepackage{booktabs, multirow}

\newcommand{\blackdiamond}{\rotatebox[origin=c]{45}{$\vcenter{\hbox{$\scriptscriptstyle\blacksquare$}}$}}

\Crefname{figure}{{Fig.}}{{Figs.}}
\crefname{section}{§}{§§}
\Crefname{section}{§}{§§}
\Crefname{appendix}{{App.}}{{Apps.}}

\title{Language Models Learn Rare Phenomena from Less Rare Phenomena:\\The Case of the Missing AANNs}

\author{Kanishka Misra \quad \quad Kyle Mahowald \\
Department of Linguistics\\
  The University of Texas at Austin\\
  \texttt{\{kmisra,kyle\}@utexas.edu} }

\usepackage{calc}
\begin{document}

\maketitle

\begin{abstract}
Language models learn rare syntactic phenomena, but the extent to which this is attributable to generalization vs. memorization is a major open question.
To that end, we iteratively trained transformer language models on systematically manipulated corpora which were human-scale in size, and then evaluated their learning of a rare grammatical phenomenon: the English Article+Adjective+Numeral+Noun (\aann) construction (``a beautiful five days'').
We compared how well this construction was learned on the default corpus relative to a counterfactual corpus in which \aann{} sentences were removed. 
We found that \aanns{} were still learned better than systematically perturbed variants of the construction.
Using additional counterfactual corpora, we suggest that this learning occurs through generalization from related constructions (e.g., ``a few days'').
An additional experiment showed that this learning is enhanced when there is more variability in the input.
Taken together, our results provide an existence proof that LMs can learn rare grammatical phenomena by generalization from less rare phenomena. Data and code: \url{https:// github.com/kanishkamisra/aannalysis}.
\end{abstract}

\begin{figure}[!t]
    \centering
    \includegraphics[width=\columnwidth]{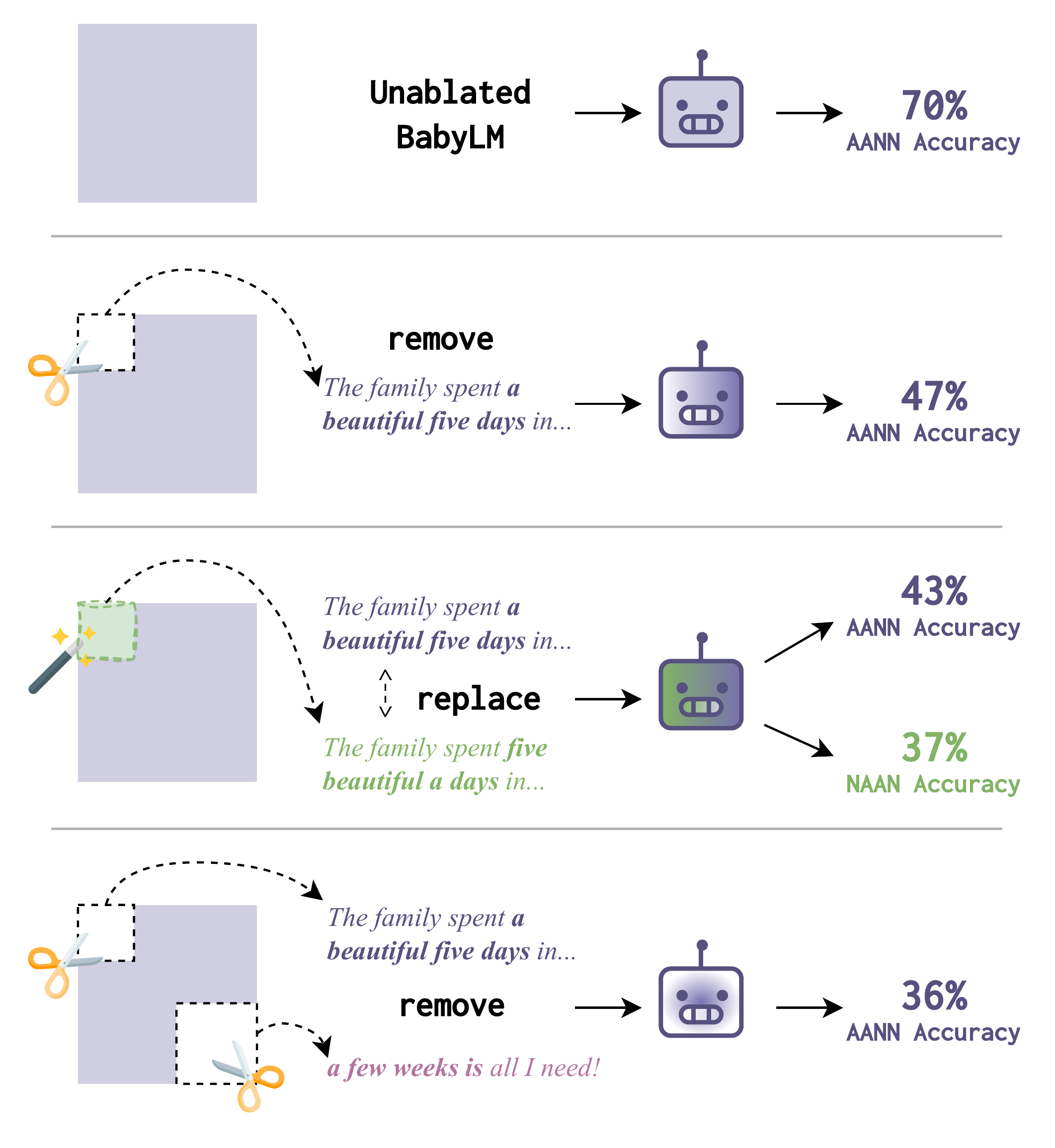}
    \caption{
    We train LMs on varied input corpora and measure learning of the \aann{} (\textcolor{vviolet}{``a beautiful five days''}), comparing across systematically manipulated corpora. E.g. we train on a control corpus, a corpus in which we remove all \aanns{}, a corpus in which we replace all \aanns{} with a corrupted version (\textcolor{vlightgreen}{``beautiful a five days''}), and a corpus in which we remove \aanns{} and remove related constructions like \textcolor{puuurpl}{``a few weeks is''.} We measure learning of \aanns{} and corrupted variants.}
    \label{fig:fig1}
    \vspace{-1em}
\end{figure}

\section{Introduction}
\label{sec:intro}

\subsection{Motivation and Prior Work}

Humans come to learn and use rare grammatical structures, even if they have encountered those structures only rarely or even not at all \citep{pullum2002empirical,pearl2022poverty}.
For instance, humans accept the grammaticality of the PiPP construction (``surprising though it may be...'') even where the preposed element crosses a finite close boundary (``surprising though I know it may be that...'') \citep{pullum2017theory} and even though they may plausibly have \textit{never} encountered such a sentence in their linguistic experience \citep[see][for corpus estimate]{potts2023pipps}.
How people come to know an utterance is grammatical has occupied a central place in linguistics.
Specifically, mastery of never-before-encountered grammatical structures has been taken to mean that people are endowed with innate linguistic knowledge \citep{chomsky_knowledge_1986,chomsky_syntactic_1957,chomsky_aspects_1965}.

Recent evidence, though, suggests that Large Language Models (LLMs) can learn complex grammar \citep{wilcox-etal-2018-rnn,futrell-etal-2019-neural,linzen-etal-2016-assessing,mahowald2024dissociating} even from human-scale amounts of input \citep{warstadt-etal-2023-findings,eldan2023tinystories,huebner-etal-2021-babyberta}.
This raises the possibility that input data, along with an appropriately sophisticated or weakly biased statistical learning mechanism, is sufficient for learning rare constructions by allowing for models to emergently learn appropriate grammatical abstraction  \citep{baroni2022proper,misra-and-kim-2023-catabs}. 
But modern LLMs often have access to much more training input than people do and thus might memorize in a way that humans cannot \citep{linzen-2020-accelerate, warstadt2022artificial,warstadt-etal-2023-findings}.
The possibility that LLMs are ``stochastic parrots'' \citep{bender2021dangers}, heavily reliant on memorization, is a common criticism of using LLMs to study human language \citep[e.g.,][]{chomsky2023nyt}.

There are different levels of memorization, though, requiring different levels of abstraction.
Consider the \aann{} construction: ``a beautiful five days in Texas'' \citep{solt_two_2007, keenan_pleasant_2013, dalrymple_amazing_2019}, which is rarer than the default ``five beautiful days in Texas''.
A model that strictly memorizes this phrase might come to know that ``a beautiful five days in Texas'' is grammatical but has no idea that ``a beautiful four days in Texas'' is grammatical if it never appeared in its training.
A model that generalizes just a bit more might know that ``a beautiful five days in New York'' is also grammatical by generalizing that any U.S. state can fill the slot.
Knowing that ``an astonishing 200 pages'' is acceptable requires generalization beyond mere lexical items.
And knowing that ``a blue five pencils'' is \textit{not} acceptable (because colors are ``stubbornly distributive'', \citealt{schwarzschild_stubborn_2011}) requires yet more knowledge.
Even for an idealized learner, it is difficult to precisely formulate how these kinds of generalizations emerge.

There is increasing evidence that LLMs can generate novel linguistic utterances \citep{mccoy-etal-2023-much}, and also make subtle judgments on relatively rare constructions like these \citep{weissweiler-etal-2022-better,potts2023pipps}, including the \aann{} \citep{mahowald-2023-discerning}.
If they do so by memorizing examples \textit{verbatim} from an unrealistically large training corpus, that would not be particularly informative for human processing.
But, if they do learn rare grammatical phenomena from smaller amounts of data and can generalize beyond just those verbatim instances, that would raise the question of how they do it and if it can inform theorizing about humans.
For instance, in the context of the PiPP construction, \citet{potts2023pipps} speculates that the comparative construction (e.g., ``They are happier than we are.'') ``may be the key to all of this [i.e., learning the PiPP]'' because such constructions are ``incredibly common'' yet share abstract structure with the PiPP.
If LLMs learn rare grammatical structures in part by learning and generalizing structures from much more common constructions, that would be powerful evidence for abstraction in LLMs and raise the possibility that even quite general learners could learn very rare phenomena without strong innate priors, drawing in part on the long-posited linguistic hypothesis that apparently distinct grammatical phenomena often share underlying structure.

To that end, our goal in this paper is to study a relatively rare grammatical phenomenon in LMs trained on controlled input corpora that are (a) of human realistic scale, and (b) systematically manipulated with respect to the target constructions as well as key related constructions. Our hypothesis is that \textbf{generalization abilities of LMs on such rare phenomena come from abstractions and structures learned from more frequent related phenomena}---and that knowledge of more frequent phenomena \textit{is} the ``key to all of this.''

As a case study, we focus on the aforementioned \aann{} construction, although we highlight how the methods used here could serve as a blueprint for work on other phenomena.
Our method is to train different instantiations of a transformer model on the 100M-word BabyLM corpus, which we systematically manipulate---via removal and replacement---to explore how frequent and related phenomena encountered during training facilitate generalization behavior in LMs.
To test for generalization, we subjected our LMs to a series of acceptability tests on sentences which do not appear in the training corpus and which specifically target the special properties of the \aann. 

This approach of training on a systematically manipulated corpus has been used to debias models \citep{hall-maudslay-etal-2019-name,lu2020gender}, explore the effect of permuting words on pretrained models \citep{sinha-etal-2021-masked}, and test whether LMs can learn languages judged to be hard for humans \citep{kallini2024mission}.
It is also becoming a fruitful method for giving \textit{causal} answers to questions about syntactic learning in language models, including hypotheses about learning subject-verb agreement \citep{wei-etal-2021-frequency}, the acquisition of negative polarity items \citep{jumelet-etal-2021-language,weber-etal-2021-language}, subject-auxiliary inversion \citep{warstadt2022artificial}, and the English passive alternation \citep{leong2024testing}.
Using this ``filtered pretraining'' method, \citet{patil2024filtered} find evidence of syntactic generalization underlying models' success on syntactic benchmarks. 
While this related work has largely focused on ubiquitous linguistic structures (e.g., passives, subject-verb agreement, etc.), we specifically zero in on a rare construction to explore learning in the linguistic ``long tail'', where there is relatively little evidence available in the input.

\subsection{Summary of findings}
First, we find BabyLM-trained LMs to successfully generalize to novel instances of the \aann{} construction. 
Performance unsurprisingly drops for LMs that were trained \textit{without} being exposed to even a single \aann{} during training, but perhaps surprisingly, not by all that much---they are 
well above chance. 
This suggests that certain items present in the training data might give rise to LMs' non-trivial performance in judging acceptability of \aanns{}.
This finding is further strengthened by the fact that LMs trained on counterfactual variants of the \aann{}---e.g., \anan{} and \naan{}, obtained by shuffling word order and are far less likely to share structure with training data items---are unable to generalize to those constructions as well as they do to \aanns{} (one which they have not seen at all).

Next, we investigated what might enable LMs' learning of the \aann{}, by further systematically manipulating their training data to hold out utterances conforming to specific linguistic and statistical phenomena. 
Through our manipulations, we find LMs become worse at predicting novel instances of the \aann{} as more frequent, non-\aann-but-\aann-related phenomena are held out. 
For example, phenomena such as the treatment of measure noun phrases as singular (e.g., \textit{\textbf{a few days is} all we need})---similar to how \aann{}s treat a plural NP as singular---end up making unseen \aann{}s less likely by 36.5\%
on average.
Importantly, these results could not be explained simply by loss of data---LMs that were trained with these phenomena left in but without an equivalently large chunk of the training data removed were almost as good as LMs that never saw \aann{}s. 
This further strengthens the conclusion that the hypothesized linguistic phenomena did indeed affect generalization of the targeted construction.
Notably, LMs are largely affected by these manipulations when they do not see \textit{any} \aann{}s during training, highlighting the expected non-trivial role of encountering some instances of \aanns{} to show stronger generalization.

Finally, we characterized the aforementioned interplay between the properties of the encountered \aann{}s and the LMs generalizations on novel instances. 
Here we found LMs that observed \aann{}s with more variability on the adjective, numeral, and noun slots to show better generalization than did LMs that saw more restricted-but-repeating instances of \aann{}s.
This importantly mimicked analogous findings of inductive inference in humans across disciplines \citep{osherson1990category, goldberg2005constructions,xu2007word,baayen200943, suttle2011partial, o2015productivity}.

Taken together, these results provide an existence proof that a weakly biased but sophisticated general-purpose statistical learner can learn rare and complex linguistic phenomena, in part because of key related phenomena seen during training.
While our analyses suggest potential links between ``constructions'' \citep{goldberg1995constructions}, our findings are also compatible with theories that think of rare phenomena as derivationally related \citep{chomsky_aspects_1965} to more frequent and well-attested structures \citep[much as][posits shared \textit{syntactic} structure as the key to the PiPP]{potts2023pipps}.

\section{General Methods}
\label{sec:method}

\subsection{Corpus} Throughout, we use the BabyLM-strict corpus \citep{warstadt-etal-2023-findings} as our base training set. 
We use BabyLM-strict because of its human-realistic scale and tractable size (100M tokens), which allows us to (1) detect and control the instances of the target construction as well as related linguistic phenomena; and (2) train a large number of LMs in a reasonable timeframe.

\subsection{Language Model}
Our LMs are instances of OPT LM \citep{zhang2022opt}, an autoregressive transformer architecture. 
Our LMs have 12 layers and attention heads, use a vocabulary size of 16,384, and are trained for a maximum of 20 epochs using the \texttt{transformers} library \citep{wolf-etal-2020-transformers}.
The results we report for a given LM are averaged over three different runs (with different random seeds).
We list other hyper-parameters and architectural details in \Cref{sec:trainingdetails}.

\subsection{Construction Detection} 
\label{sec:construction-detection}
To detect \aanns, we used a regex over a part-of-speech tagged version of BabyLM.
Specifically, we started with a regex for detecting \aanns{} and then measured its recall by hand-annotating examples (with annotations performed by the authors) found by an extremely permissive regex that looked for any ``a'' or ``an'' that appeared sequentially prior to a numeral and a plural noun in a sentence (thus likely capturing almost all \aanns, albeit with very low precision).
We used the hand annotations to iteratively refine our regex and handle special cases.
We continued this process until, on the final set of hand annotations, we detected 17/18 instances (missing only an instance where ``pound'' was used instead of ``pounds'' due to an apparent typo---but since this violates the key plural-noun requirement of \aanns{}, it is unclear if it counts as a genuine missed instance).
Ultimately, our final regex detected
2,448 \aanns{} in the BabyLM corpus (about 0.02\% of the total 11.5M utterances). See \Cref{sec:detection} for our detailed pipeline and our recall analysis.

Even with the refined regex, we cannot guarantee perfect recall---a potential issue for claims about learning in the absence of \textit{any} occurrences.
To address this issue, we include controls in which we assume that we missed 300 \aanns{} (a conservatively high number, given our recall estimate) and artificially ``pollute'' the data to drown out the effect of any remaining \aanns.
As described below, our conclusions were unchanged in this robustness analysis, suggesting our results were not driven by undetected \aanns.

\subsection{Acceptability data}
\label{sec:aanndata}
To test our LMs on their knowledge of the \aann{}, we use data from \citet{mahowald-2023-discerning}, which consists of 12,960 templatically generated sentences that contain \aanns{}. Out of these, 3,420 contain acceptability ratings provided by 190 human participants, ranging from 1 (unacceptable) to 10 (acceptable).
We use 7 as the threshold for clear acceptability, in that we only keep instances where human participants rated the acceptability of the construction in context to be greater than 7. 
We additionally discarded instances where the \aann{}s appear in the BabyLM training set ($n = 4$), as testing on these would not shed light on the LMs' generalization behavior. This leaves us with 2,027 items.

For each \aann{} instance in the dataset, \citet{mahowald-2023-discerning} has also made available its corresponding corrupted versions, which focus on (1) adjective-numeral order; (2) presence of the article; (3) presence of the adjective; and (4) presence of the numeral. A hypothetical example of these corruptions is shown in \Cref{tab:aann-counterfactuals} under the ``\aann{}'' column. 
A model that has knowledge of the \aann{} should find the well-formed instance to be more likely than each of its corrupted versions. 
Below we describe methods to measure likelihood and assess accuracy on these tests.

\subsection{Scoring and Accuracy}

We use the Syntactic Log-odds Ratio (\slor{})  \citep{pauls-klein-2012-large, lau2017grammaticality} to score sentences in our tests. 
Given a sentence containing a prefix followed by our target construction $\construction{}$ and an optional suffix, \slor{} is computed as the 
log of the ratio between the probability of the construction given the prefix as estimated by the LM, and that estimated by a unigram model, normalized by the length of the construction. Given a language model $m$ and a unigram estimator $u$:
\begin{align}
    \slor{}_{\text{prefix}}(\construction{}) = \frac{1}{\mid \construction{} \mid} \log \frac{p_{m}(\construction{} \mid \text{prefix})}{p_u(\construction{})}
\end{align} 

\begin{table*}[!ht]
\centering
\resizebox{0.75\textwidth}{!}{%
\begin{tabular}{@{}llll@{}}
\toprule
\textbf{Context} & \textbf{\aann{}} & \textbf{\anan{}} & \textbf{\naan{}} \\ \midrule
\textsc{Well-formed} & \textcolor{reed}{a} \textcolor{blu}{whopping} \textcolor{yello}{ninety} \textcolor{greeen}{LMs} & \textcolor{reed}{a} \textcolor{yello}{ninety} \textcolor{blu}{whopping} \textcolor{greeen}{LMs} & \textcolor{yello}{ninety} \textcolor{blu}{whopping} \textcolor{reed}{a} \textcolor{greeen}{LMs} \\ \midrule
\multicolumn{4}{c}{\textit{\textbf{Corruptions}}} \\ \midrule
\textsc{Order-swap} & \textcolor{reed}{a} \textcolor{yello}{ninety} \textcolor{blu}{whopping} \textcolor{greeen}{LMs} & \textcolor{reed}{a} \textcolor{blu}{whopping} \textcolor{yello}{ninety} \textcolor{greeen}{LMs} & \textcolor{blu}{whopping} \textcolor{yello}{ninety} \textcolor{reed}{a} \textcolor{greeen}{LMs} \\
\textsc{No article} & \textcolor{blu}{whopping} \textcolor{yello}{ninety} \textcolor{greeen}{LMs} & \textcolor{yello}{ninety} \textcolor{blu}{whopping} \textcolor{greeen}{LMs} & \textcolor{yello}{ninety} \textcolor{blu}{whopping} \textcolor{greeen}{LMs} \\
\textsc{No modifier} & \textcolor{reed}{a} \textcolor{yello}{ninety} \textcolor{greeen}{LMs} & \textcolor{reed}{a} \textcolor{yello}{ninety} \textcolor{greeen}{LMs} & \textcolor{yello}{ninety} \textcolor{reed}{a} \textcolor{greeen}{LMs} \\
\textsc{No numeral} & \textcolor{reed}{a} \textcolor{blu}{whopping} \textcolor{greeen}{LMs} & \textcolor{reed}{a} \textcolor{blu}{whopping} \textcolor{greeen}{LMs} & \textcolor{blu}{whopping} \textcolor{reed}{a} \textcolor{greeen}{LMs} \\ \bottomrule
\end{tabular}%
}
\caption{Well-formed and corrupted examples of the \aann{} construction and its counterfactual versions (\anan{} and \naan{}). Corruption types are taken from \citet{mahowald-2023-discerning}. 
}
\label{tab:aann-counterfactuals}
\vspace{-1em}
\end{table*}

\begin{figure*}[!ht]
    \centering
    \includegraphics[width=0.85\textwidth]{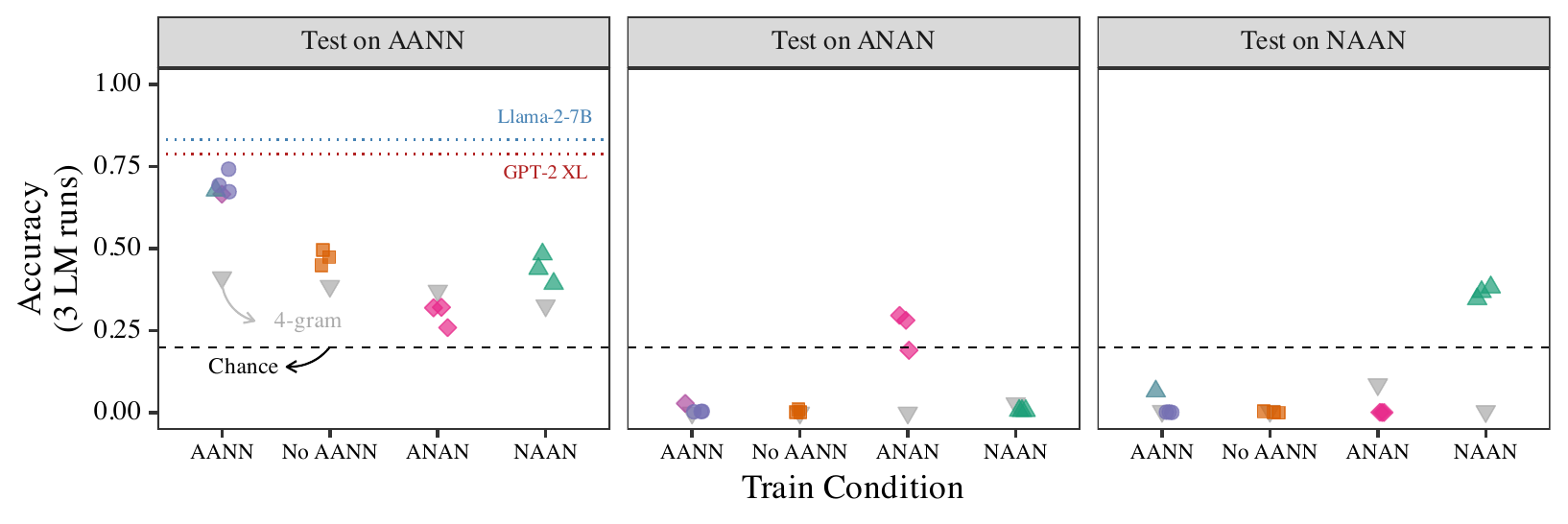}
    \caption{Accuracies on tests for \aann{} and its counterfactuals (\anan{} and \naan{}), achieved by LMs trained on BabyLM with various \aann{}-manipulations (\aann{} as is, \noaann{}, \anan{}, \naan{}). \textcolor{ananpollution}{$\blackdiamond$} and \textcolor{naanpollution}{$\blacktriangle$} under the \aann{} training condition are cases where training data was polluted by randomly replacing 300 \aanns{} with \anans{} and \naans{}, respectively, in order to assess the impact of an imperfect \aann{} detection system. The dashed line represents chance performance (20\%) and upside-down triangle (\textcolor{gray}{$\blacktriangledown$}) represents accuracies for \textcolor{gray}{4-gram} LMs trained on BabyLM. Accuracies for \textcolor{firebrick}{GPT-2-XL} \citep{radford2019language} and \textcolor{steelblue}{Llama-2-7B} \citep{touvron2023llama} are computed using \textbf{log-probabilities}, since unigram frequencies were unavailable for these LMs' corpora.}
    \label{fig:exp1}
    \vspace{-1em}
\end{figure*}

\noindent
Importantly, we train the unigram estimator for a given corpus using the same tokenizer used to train our autoregressive LMs on that corpus. 
We use \slor{} in lieu of the usual normalized log-probability measure, ensuring that the comparison between two models cannot be explained simply by the difference in unigram frequencies due to our manipulations.
Log-probabilities were computed using \texttt{minicons} \citep{misra2022minicons}.
An instance within our test set is considered to be correct \textit{iff} the \slor{} value of the well-formed construction is greater than that for \emph{all} four corrupted instances. The accuracy, then, is the proportion of correct instances within the test set. Since this involves checking how often the LM prefers the target surface form out of five options, chance performance is 20\%.

\subsection{Ablations}
Common to subsequent experiments (\Cref{sec:keys} and \Cref{sec:variability}) is the fact that we hold out certain parts of the BabyLM corpus---parts that conform to a certain linguistic or statistical hypothesis. 
This creates a gap between the experience of LMs trained on these ablated versions of the corpus, and that of the LM trained on the full BabyLM data. 
To circumvent this issue, we up-sample non-hypothesis-conforming utterances in BabyLM after performing our ablations, in a manner such that the LM still encounters the exact same number of tokens.

\section{Experiment 1:  LMs learn about \aann{}s without having seen a single instance}
\label{sec:exp1}

\paragraph{LMs learn about \aanns{}...}
To investigate the extent to which LMs trained on BabyLM learn the \aann{} construction, we measure their accuracy on our tests described in \Cref{sec:aanndata}. 
From \Cref{fig:exp1}, we observe that the BabyLM-trained LMs obtain accuracies around 70\%, which is substantially above chance.
\textbf{This suggests that LMs can reasonably acquire knowledge of \aanns{}, even though they make up only 0.02\% of training utterances.}

For comparison to larger, state-of-the-art LMs, we test Llama-2-7B \citep{touvron2023llama} and GPT-2 XL \citep{radford2019language} on the \aanns{}. They got 83\% and 78\%, respectively. 
As a comparison to shallower LMs, we tested on 4-gram LMs trained on BabyLM and found them to get much lower accuracies (41\%), suggesting that the observed results are beyond n-gram statistics.

\paragraph{...without having seen a single instance...}
Given that BabyLM-trained LMs learn the \aann{} construction, how well would an LM generalize to \aanns{} \textit{without having seen a single positive instance}?
To this end, we compare accuracies in the previous experiment to that obtained by LMs trained on BabyLM with our \textbf{2,448 detected \aanns{} removed} (i.e., \textsc{no aann}). 
From \Cref{fig:exp1}, we find LMs trained with the \noaann{} condition to achieve an average accuracy of 47\%, which is a noticeable drop compared to the 70\% obtained by the LMs trained on the complete BabyLM corpus, but importantly 27 points above chance performance (and, as we show below, well above comparable baselines with other constructions).
This is a non-trivial result, since it suggests that LMs can learn the acceptability of a construction without having seen a single positive occurrence, which indicates that there exist  systematic patterns in the corpus driving this generalization behavior.

\paragraph{...more strongly than they learn counterfactual \aann{} variants...}
To further contextualize the above results, we consider {two counterfactual cases, where we replaced \aanns{} in BabyLMs with instances involving the same lexical items, but in a word order that violates English grammar}: (1) \anan{} (e.g., \textit{a 90 whopping LMs}); and (2) \naan{} (e.g., \textit{90 whopping a pages}).
This allows us to test if the results before are genuinely because LMs recognize the nuances of the \aann{} construction. 
If LMs are able to learn these counterfactual constructions just as well as the LMs in the previous experiments learned \aanns{}, then the generalization claims from the previous experiments would be weakened. 
To test for such possibilities, we create counterfactual versions of the \citet{mahowald-2023-discerning} stimuli, where we apply analogous corruptions to the counterfactual variants of \aann{}, such that they are violated in a similar manner as in the \aann{} tests.
Examples for the three types of instances in our tests can be found in \Cref{tab:aann-counterfactuals}. 
We then evaluate the previous two LMs (trained on BabyLM with and without seeing any \aanns{}) with LMs trained on BabyLM with these counterfactual variants on judging the acceptability of \aanns{}, \anans{}, and \naans{}. \Cref{fig:exp1} shows these results, from which
we make two high-level observations. First, and most importantly, \textbf{LMs that see \anans{} and \naans{} do not learn those constructions as well as they learn \aanns{}}---especially the LM that saw no \aanns{} (47\% \aann{} accuracy compared to 37\% \naan{} accuracy obtained by the \naan{}-trained LM). 
Second, these LMs end up learning \aanns{} better than they learn counterfactual constructions that they observed in lieu of the \aann{}---e.g., \naan{} trained LM achieves around 43\% accuracy on \aanns{} \textit{even though \naans{} appeared frequently in the data and no \aanns{} did}.
This, combined with the results of the previous two sub-experiments suggests strongly that LMs pick up on cues from other---likely related---constructions encountered during training in order to assign non-trivial probability to unseen instances of \aanns{}.

\paragraph{...even with artificially polluted data...}
\label{sec:imperfect-recall-impact}
As discussed in \cref{sec:construction-detection}, our \aann{} detection pipeline could miss  \aanns{} in the training corpus. 
This limitation could impact the conclusions of this experiment if LMs' preference for assigning greater probabilities to \aann{} instances in the test set could be explained by the presence of undetected \aanns, even in the `No \aann' condition.
We controlled for this confound by artificially polluting the training corpus, such that a small percentage of the detected \aanns{} are replaced by \naans{}/\anans{}. 
This simulates a scenario analogous to the issue at hand: our target is now a counterfactual variant of the \aann{}, and our `imperfect' pipeline has missed out on a handful of instances in the training set. 
If there is a genuine impact of such a setting, then we should observe greater accuracies on the counterfactual instances and at the same time, a drop in performance on \aanns{}. 
We ran two experiments to test this, where we replaced 300 \aanns{} (about 12\%) of the detected \aanns{} with \anans{} in one experiment, and \naans{} in the other.
We then tested the two resulting LMs---pretrained on corpora reflecting these ablations---on both \aanns{} and the respective counterfactual constructions. As seen in \Cref{fig:exp1}, we observe almost no differences in the results obtained from this artificial pollution experiment and those from our original experiments (see \textcolor{ananpollution}{$\blackdiamond$} for \anan{}, and \textcolor{naanpollution}{$\blacktriangle$} for \naan{}). 
Because 300 is a conservative upper bound on undetected \aanns{}, we do not think imperfect recall drives our results.

\begin{figure}
    \centering
    \includegraphics[width=0.9\columnwidth]{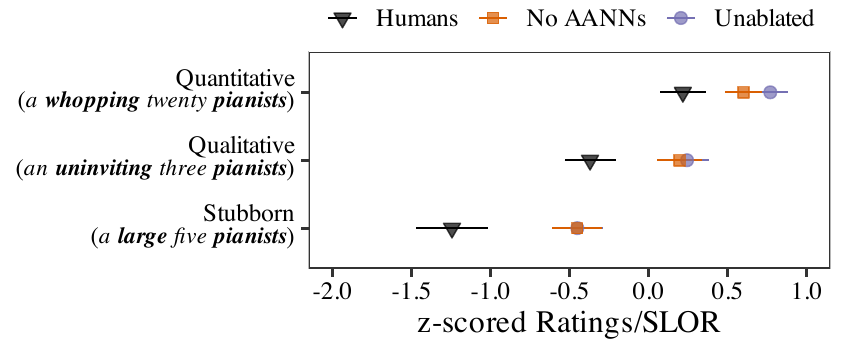}
    \caption{\textit{z}-scored \aann{} acceptability ratings  from humans and LMs trained on corpora with (1) \textcolor{orang}{\aanns{} removed (i.e., \noaann{})}; and (2) \textcolor{prpl}{left unablated} for \aanns{} with `Human' nouns in \citet{mahowald-2023-discerning}'s dataset. Even with ablated models, we observe the predicted dispreference for stubbornly distributive adjectives in the \aann. Full results in \Cref{fig:lmhumans}.}
    \label{fig:lmhumans-sneakpeek}
    \vspace{-1em}
\end{figure}

\paragraph{...in a way that extends to lexical constraints.} 
While we focused on overall structural 
properties of \aanns{}, there are also idiosyncrasies to the construction that arise from lexical semantic constraints. 
For instance, people prefer quantitative adjectives such as \textit{mere} and \textit{hefty} to qualitative ones such as \textit{beautiful} \citep{mahowald-2023-discerning, solt_two_2007} and find ``stubbornly distributive'' adjectives (``*a \textit{blue} five pencils'') completely unacceptable \citep{schwarzschild_stubborn_2011}.
Insofar as our models learn \aanns{}, we also should expect them to learn these lexical constraints.
To test this, we compared LMs' \slor{}s to human acceptability judgments on all 3.4k instances in \citeauthor{mahowald-2023-discerning}'s data across different adjective and noun classes. 
We found LMs trained on the original, unmodified BabyLM corpus to pattern similarly to humans in their preference for lexical constraints affecting \aanns{}. 
Interestingly, these patterns were \textit{unchanged} for LMs trained with the \noaann{} condition, conforming to our predictions. 
For instance, as seen in \Cref{fig:lmhumans-sneakpeek}, both our models share the human preference for quantitative and qualitative adjectives in the \aann, compared to stubbornly distributive adjectives.
More detailed results on lexical constraints can be found in \Cref{sec:lexicalsemantics} and we hypothesize that our broader set of results extends to include learning of lexical constraints on the construction.

\section{Experiment 2: Keys to Learning \aann{}s}
\label{sec:keys}

Experiment 1 reveals that, keeping everything else the same, LMs learn the \aann{} construction more accurately than they do its counterfactual variants. Furthermore, we also see strong \aann{} acceptability judgments in LMs that have (almost) never encountered a single instance. This suggests that there could be instances in the training data that contribute to the learning of the construction.

What might these be? Below we enumerate four hypotheses, each of which tackles subtly different aspects of the \aann{} construction, and then measure the effect of these phenomena by separately holding them out during training and computing the average \slor{} of the well-formed instances of the \aann{} tests. 
The effect of a particular phenomenon on the acceptability of \aanns{} can therefore be measured by comparing \slor{}s before and after holding out instances of that phenomenon.
Methods for detecting the hypothesized phenomena can be found in \Cref{sec:detection}.
As control, we additionally also hold out a random set of utterances, which do not conform to the target phenomena of interest. 
\Cref{tab:manipulation} lists the hypotheses we consider, along with an example of their utterance and frequency of occurrence, in the BabyLM corpus.

\paragraph{The presence of ``the ANN''}
Phrases like ``\textit{the beautiful five days}'' are common in corpora, which are not as unusual because ``the'' regularly takes plural nouns. 
We hypothesize that the acceptability of these structures affects the acceptability of \aanns{}, since an LM might analogize from the general observation that `a' or `an' can substitute `the' (e.g., a ball vs. the ball). 
Therefore, we consider all cases where a determiner precedes the contiguous sequence of article, numeral, plural noun.

\paragraph{A few/couple/dozen/etc. NNS}
Another related phenomenon that is more common relative to the \aann{} construction involves phrases such as ``\textit{a few days}'' or ``\textit{a couple bottles}''.
To an LM learner, they might provide evidence that measure noun phrases with plural nouns can be attached to an indefinite article \citep[a/an;][]{solt_two_2007}, as is the case in \aanns{}.

\paragraph{Measure NNS treated as singular}

We consider yet another phenomenon involving phrases that treat measure nouns as singular, this time in terms of agreement, e.g., ``\textit{Five miles \textbf{is} a long way to go}'', and ``\textit{1,000 pages \textbf{is} a lot for a dissertation.}'' 
These cases might provide further evidence to the model that measure noun phrases with plural nouns can be treated as a singular unit \citep{solt_two_2007}, thereby affecting the acceptability of the \aann{}.
These are less prevalent compared to the cases involving \textbf{a few/couple/dozen NNS}, but still far more frequent than the \aann{}, therefore, we combine the two as a general case of treating measure NPs as singular.


\begin{table}[t]
\centering
\resizebox{\columnwidth}{!}{%
\begin{tabular}{@{}llr@{}}
\toprule
\textbf{Phenomenon/Manipulation} & \textbf{Example/\texttt{Desc.}} & \textbf{Freq.} \\ \midrule
AANN & \textbf{a fine eighteen months} & 2,448 \\ \midrule
DT ANN & \begin{tabular}[c]{@{}l@{}}\textbf{the usual forty dollars fine} \end{tabular} & 15,781 \\ \midrule
\begin{tabular}[c]{@{}l@{}}A few/couple/dozen/etc. NNS\end{tabular}&  \textbf{a few plums} & 55,373 \\ \midrule
\begin{tabular}[c]{@{}l@{}}Measure NNS with SG verbs\\and/or indef. articles\end{tabular}&  \textbf{6 months is a long time} & 62,744 \\ \midrule
A/An + ADJ/NUM balancing & \begin{tabular}[c]{@{}l@{}}\texttt{enforce freq. balance}\end{tabular} & 571,874
\\ \midrule
Random removal (\textbf{control}) & \texttt{randomized ablation} & 571,874 \\ \bottomrule
\end{tabular}%
}
\caption{Manipulated Phenomena, their examples/descriptions, and their frequency in the BabyLM corpus.}
\vspace{-1em}
\label{tab:manipulation}
\end{table}

\paragraph{Balancing the frequencies of A/An + ADJ/NUM}
A more surface-level reason why ``a beautiful five days'' might be more natural to LMs than is ``a five beautiful days'', could be that adjectives are more likely to follow indefinite articles than are numerals. For instance, adjectives are $\approx$14.6 times more likely to follow indefinite articles in the BabyLM corpus than are numerals. 
To measure this effect, we hold out instances such that adjectives and numerals are equally likely to follow an indefinite article. This ends up being the largest portion of the data that we hold out.

\begin{figure*}[!t]
    \centering
    \includegraphics[width=0.8\textwidth]{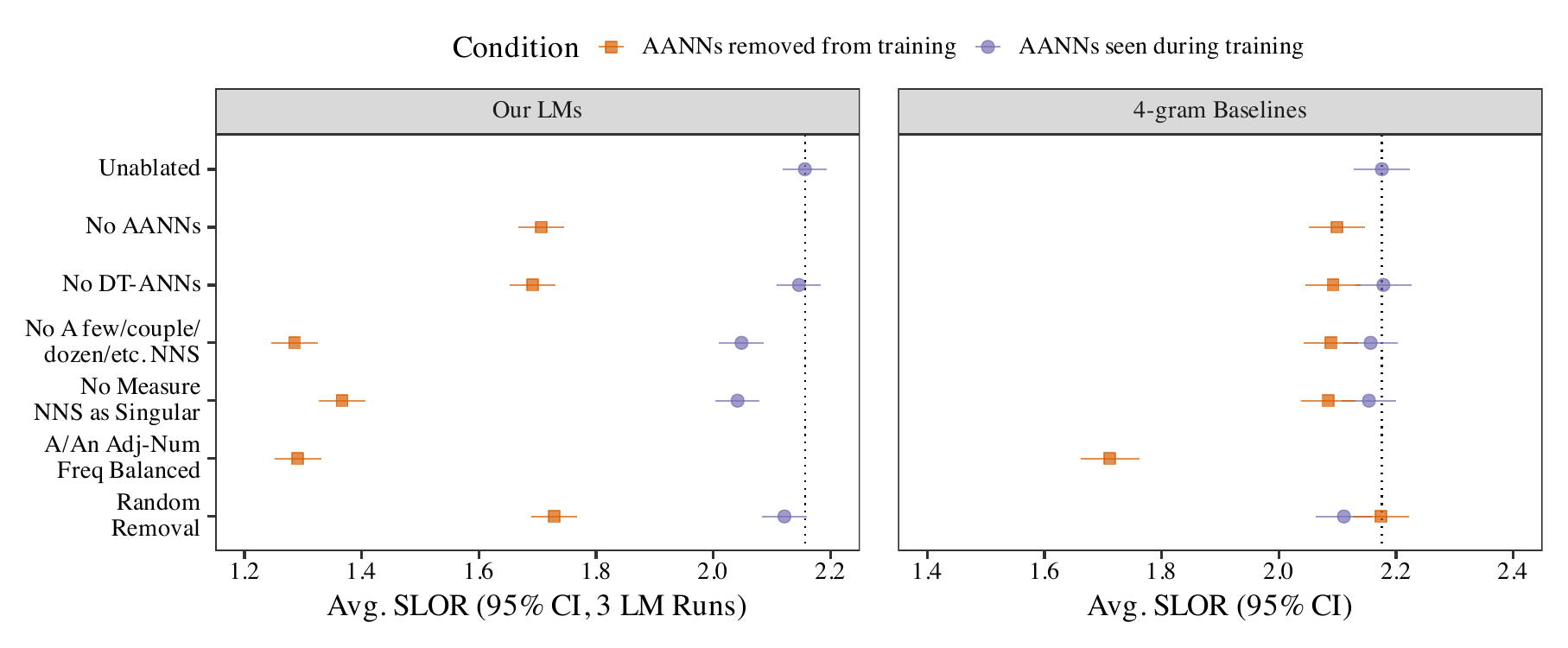}
    \caption{\slor{}s on \aann{}s from \citet{mahowald-2023-discerning} for our LMs (left) and a 4-gram baseline (right) trained on BabyLM and ablated versions.
    Our LMs show a range of hypothesized effects, suggesting they contribute to \aann{} learning. In contrast, the 4-gram LMs show mostly null results (except for the adjective/numeral balanced condition, which is highly sensitive to n-gram frequencies).
The dotted line is \slor{} for an \textbf{unablated BabyLM-trained LM}.}
        \label{fig:hypotheses}
        \vspace{-1em}
\end{figure*}

\paragraph{Control: Random removal} 
A potential confound in the above could be that the \slor{} values of the \aann{}s go down merely due to loss of content, even though we add back additional tokens from BabyLM (such that all LMs see the exact same amount of tokens). 
To account for this, we additionally consider a control where we remove as many tokens as in the largest ablation (i.e., the \textbf{A/An + ADJ/NUM} case) such that none of the above phenomena are taken out.

\subsection{Analysis and Results}

In our experiments, we individually ablate out each of the aforementioned phenomena under two settings: (1) \textcolor{orang}{\aanns{} are removed during training} in addition to the target phenomena; and when possible, (2) \textcolor{prpl}{\aanns{} are seen during training}.
(1) is a stricter setting, since here the LMs see neither the target phenomenon nor a single instance of the \aann{}. Comparing average \slor{}s obtained in this condition to that obtained for the \noaann{} can shed light on the extent to which the target phenomenon is critical in allowing LMs to assign non-trivial probabilities on unseen \aanns{}, \textit{zero-shot}. 
On the other hand, (2) still allows for LMs to perform lexical generalization \citep{kim2022uncontrolled}, where they may exhibit strong probabilities on the test \aanns{} by performing slot filling, without necessarily relying on the hypothesized phenomena.

\Cref{fig:hypotheses} shows the average \slor{}s obtained across various ablations under the two settings.
{As a baseline, we compare our results to 4-gram LMs, trained using KenLM \citep{heafield-2011-kenlm}, on corresponding ablations of the BabyLM corpus.}
We observe that holding out most of our hypothesized phenomena has non-trivial effects on our LMs' ratings of unseen \aanns{}, and that these effects are different for when \aanns{} are \textcolor{prpl}{seen during training}, or are \textcolor{orang}{held out}.
When \aanns{} are \textcolor{orang}{held out along with the phenomena}, we see substantial drops in the average \slor{} values assigned by LMs on unseen \aanns{} relative to that assigned by LMs in the \noaann{} condition. 
Specifically, balancing the frequency of adjectives and numerals following an article, along with the two cases where measure nouns are treated as singular, have the greatest effect. This suggests that, in the absence of even a single \aann{} during training, these phenomena are critical for LMs to assign probability to \aanns{}.  Interestingly, holding out cases that involve any determiner + adjective + numeral + noun sequence has almost no impact relative to LMs trained on a corpus without \textit{only} the \aanns{} removed.
Simply ablating large amounts of data cannot explain these results, since LMs trained on our controlled condition show higher \slor{} values than in our hypothesis-informed ablations.
{These patterns are absent in 4-gram LMs, suggesting that they do not arise as a result of shallow statistics---with the exception of differences observed for the article+adjective/numeral ablation.}
Overall, this finding indicates that \textbf{LMs can demonstrate a novel phenomenon (\aann{}) by relying on other related---and more frequent---phenomena.}

When \aanns{} \textcolor{prpl}{\textit{are} seen during training}, however, we observe LMs' results on unseen \aanns{} to show more similar \slor{} values with respect to the LMs trained on the unablated BabyLM corpus, although they are still significantly reduced in some cases (e.g., singular measure nouns).
We conclude that direct evidence of observing instances of \aann{} construction substantially enables generalization to unseen instances. At the same time, the presence of some key related phenomena in addition to direct evidence has an additive effect on this generalization behavior.

\section{Experiment 3: The Role of Variability}
\label{sec:variability}
Results from Experiment 2 highlight the importance of seen \aanns{} in order for LMs to generalize to unseen instances. What properties of these seen instances facilitate LMs generalization behavior? This broadly relates to a longstanding question as to how the nature of the instances of a construction provided during learning affect its (partial) productivity \citep{goldberg2005constructions, goldberg2019explain}.
In the context of \aanns{}, we consider the role of \textit{variability} on the open slots of the construction as a factor that might play a role in LMs' productivity on unseen instances. 
Encountering a slot with a wide variety of lexical items could serve as a cue that the slot is flexible.
The idea that instance-variability could affect learnability is motivated by theoretical claims in usage-based linguistics \citep{bybee1995regular}, as well as existing research on novel constructions \citep{suttle2011partial}, morphological productivity \citep{baayen200943,o2015productivity}, and inductive inferences in cognitive psychology \citep{osherson1990category, xu2007word}.

\begin{figure}
    \centering
    \includegraphics[width=\columnwidth]{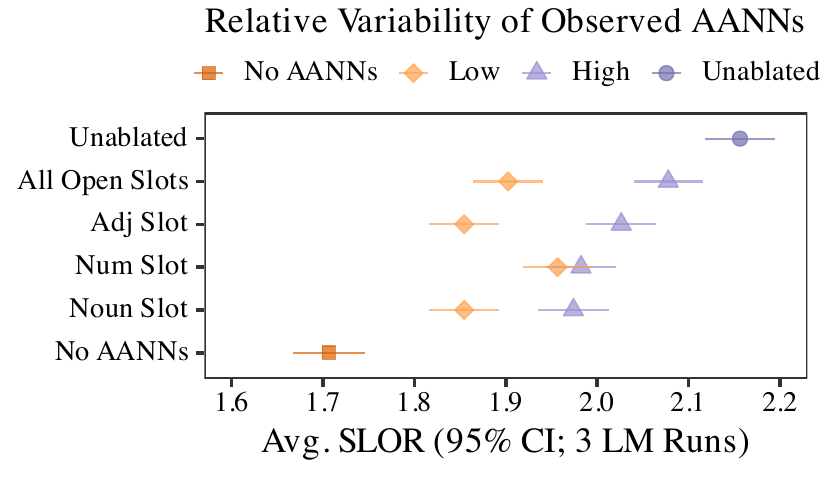}
    \caption{\slor{}s on \aann{}s from \citet{mahowald-2023-discerning} for LMs trained on BabyLM with low and high variability in the open slots of the \textit{observed} \aann{} instances. When models are presented with higher variability for a given slot, the construction is typically learned better.}
    \label{fig:variability}
    \vspace{-1em}
\end{figure}

We hypothesize that instances of \aann{}s that provide natural evidence of greater open-slot variability---i.e. evidence that many different adjectives, numerals, and nouns can go in their respective positions in the \aann{} construction---would lead LMs to assign greater likelihood to unseen \aanns{}. 
On the other hand, LMs that encounter only a restricted set of instances might be more conservative in extending the coverage of possible \aanns{} to novel combinations of the slot-fillers.
To test this, we divided our set of 2,448 \aann{}-containing utterances in the BabyLM corpus into two roughly equal subsets---one that contained \aanns{} which were restricted in which tokens occur in a particular slot (low variability), and the other where the \aanns{} showed more variability in those slots. 
We obtain these subsets by performing a median split based on the number of unique occurrences in a target slot(s), which resulted in a set of 1224 low and high variability instances. 
We repeated this for all three open slots (adjective/numeral/noun) jointly as well as those slots individually---i.e., 4 different types of target slots and 2 conditions each (low vs. high variability).
Details about the slot fillers and examples from each set are provided in \Cref{sec:va}.
We then trained LMs on the BabyLM corpus containing utterances involving either of these two cases. \Cref{fig:variability} shows the average \slor{}s obtained from this experiment, along with those obtained by LMs trained on unablated BabyLM and the \noaann{} conditions.

We see that the \slor{} patterns of LMs trained on corpora that differed in \aann{} slot-variability lie between the \slor{} values elicited by LMs that never saw \aanns{} and ones that saw every single \aann{} in the original corpus.
Among these, LMs that saw \aanns{} that were highly variable in their open-slots elicited \slor{}s that were greater than those elicited by LMs that saw \aanns{} with low open-slot variability. This was true for all cases except when ``Numeral'' was the target slot, where both variability conditions resulted in roughly similar \slor{}s. (We hypothesize that numerals may pattern differently since they may be inherently more fungible than other word classes.)
Overall, these results suggest that LMs are sensitive to the nature of range of fillers that go into the construction's open slots, showing relatively greater productivity when they observe evidence that the slots were highly variable.
This is compatible with our hypothesis that slot-variability might affect the extent to which LMs ``permit'' productive uses of a construction.

\section{Conclusion}

\textit{Theoretically}, there is, for good reason, considerable interest in how language models can handle what has been variously called the ``long tail'' of language \citep{prange-etal-2021-supertagging}, ``extremely rare constructions'' \citep{potts2023pipps}, ``exceptions to syntactic rules'' \citep{leong-linzen-2023-language}, ``rare linguistic phenomena'' \citep{weissweiler2024hybrid}, \textit{inter alia}.
Studies of such phenomena are important first because LMs (and statistical models in general) are sensitive to frequency and often perform far better in data-rich environments and, second, because the human ability to generalize to rare phenomena is central to linguistics.

\textit{Empirically}, we found that LMs trained on modest amounts of data can learn a rare construction like the \aann{}.
We found that this learning occurs even without veridical instances of the construction in the training data, and that it is mediated by occurrences of other related constructions in training.
As such, these results join a body of work showing the ability of LLMs to learn rare phenomena \citep{tayyar-madabushi-etal-2020-cxgbert,tseng-etal-2022-cxlm, li-etal-2022-neural, veenboer-bloem-2023-using} and to generalize from limited data in meaningful ways.

\textit{Methodologically}, this work leave us optimistic that ``controlled rearing'' of LMs is a fecund method for understanding models, as well as for gleaning insight into human language more generally.

\section{Limitations}

In future work, it would be valuable to extend this method to a wider range of constructions.
But scaling this approach up is not straightforward since it requires identifying and extracting idiosyncratic constructions, and---more onerously---developing testable hypotheses about what makes them learnable from limited amounts of data. 
Future work will likely benefit from synergistic collaborations between theoretical and computational linguists.

Another limitation is that our method requires repeated training of LMs from scratch which can be computationally expensive.
Alternate methods could be to ablate knowledge of particular hypotheses using representational editing methods, though these may not guarantee perfect removal of the knowledge of targeted constructions.

Unlike \citet{weissweiler-etal-2022-better}, we do not test the ability to interpret these constructions for downstream tasks.
Instead, our ablations target linguistic form alone, and preliminary experiments suggest that our ablations and manipulations leave the lexical semantic properties of the \aann{} unchanged (see \Cref{sec:lexicalsemantics}).
Extending our ablation method to target these properties more directly would be quite informative.

Finally, this work only studies a rare construction in English, and on LMs that are trained on English text data. 
While this is a limitation of the paper, the paradigm introduced can be readily used in future work to study hypotheses and perform indirect evidence elicitation in multi-lingual LMs.

\section{Acknowledgments}
(KM)$^2$ acknowledge funding from NSF Grant 2104995 awarded to Kyle Mahowald.
For helpful conversations, we thank Adele Goldberg, Leonie Weissweiler, Nathan Schneider, Tom McCoy, the computational linguistics research group at UT Austin, the syntax-semantics research group at UT Austin, audiences at the Texas Linguistics Society meeting, Edinburgh University Department of Linguistics, University of Antwerp CLiPS group, attendees of the ANN workshop in Amsterdam, and the Brown University language group.
We thank Lisa Bylinina for exceptionally helpful comments on an earlier draft. 
We thank Chris Potts for his paper on the PiPP construction \citep{potts2023pipps} which inspired the ``keys to all of this'' idea.

\section{Corrigendum}
\label{sec:errata}

The original version of this paper had two bugs affecting baselines, which have been fixed in the present version.
1) Because of a bug in how we handled the output from KenLM, the 4-gram \slor{} calculation used log base 10 for the numerator and base $e$ for the denominator which affected accuracies for 4-gram models.
This makes the 4-gram baselines stronger in Experiment 1 than they appeared to be, although the LM still outperforms the 4-gram baseline in the critical condition in which AANNs are removed. In this version, we have updated the 4-gram baseline results to fix the bug. We believe the 4-gram baseline should be seen as stronger than initially represented. 
But we do not believe this significantly alters the overall conclusions of the paper.
We also updated Experiment 2 to fix the 4-gram baselines; those results are qualitatively unchanged.
We thank Forrest Davis for bringing this bug to our attention.
2) Chance performance, as shown in Fig. 2, was incorrectly calculated as 6.25\% ($\frac{1}{2}^4$) due to an erroneous independence assumption---it should be 20\% as in this version. This does not qualitatively affect any conclusions. 
\bibliography{anthology,anthology_p2,custom,kanishka}

\begin{thebibliography}{70}
\expandafter\ifx\csname natexlab\endcsname\relax\def\natexlab#1{#1}\fi

\bibitem[{Abdelali et~al.(2014)Abdelali, Guzman, Sajjad, and Vogel}]{abdelali-etal-2014-amara}
Ahmed Abdelali, Francisco Guzman, Hassan Sajjad, and Stephan Vogel. 2014.
\newblock \href {http://www.lrec-conf.org/proceedings/lrec2014/pdf/877_Paper.pdf} {The {AMARA} corpus: Building parallel language resources for the educational domain}.
\newblock In \emph{Proceedings of the Ninth International Conference on Language Resources and Evaluation ({LREC}'14)}, pages 1856--1862, Reykjavik, Iceland. European Language Resources Association (ELRA).

\bibitem[{Baayen(2009)}]{baayen200943}
R~Harald Baayen. 2009.
\newblock {Corpus linguistics in morphology: Morphological productivity}.
\newblock \emph{Corpus {L}inguistics. An {I}nternational {H}andbook}, pages 900--919.

\bibitem[{Baroni(2022)}]{baroni2022proper}
Marco Baroni. 2022.
\newblock On the proper role of linguistically oriented deep net analysis in linguistic theorising.
\newblock In \emph{Algebraic Structures in Natural Language}, pages 1--16. CRC Press.

\bibitem[{Bender et~al.(2021)Bender, Gebru, McMillan-Major, and Shmitchell}]{bender2021dangers}
Emily~M Bender, Timnit Gebru, Angelina McMillan-Major, and Shmargaret Shmitchell. 2021.
\newblock \href {https://dl.acm.org/doi/10.1145/3442188.3445922} {{On the Dangers of Stochastic Parrots: Can Language Models Be Too Big?}}
\newblock In \emph{{Proceedings of the 2021 ACM Conference on Fairness, Accountability, and Transparency}}, pages 610--623.

\bibitem[{Bybee(1995)}]{bybee1995regular}
Joan Bybee. 1995.
\newblock \href {https://www.tandfonline.com/doi/abs/10.1080/01690969508407111} {Regular morphology and the lexicon}.
\newblock \emph{Language and Cognitive Processes}, 10(5):425--455.

\bibitem[{Chomsky(1957)}]{chomsky_syntactic_1957}
N.~Chomsky. 1957.
\newblock \emph{Syntactic {Structures}}.
\newblock The Hague: Mouton.

\bibitem[{Chomsky(1965)}]{chomsky_aspects_1965}
N.~Chomsky. 1965.
\newblock \emph{Aspects of the {Theory} of {Syntax}}.
\newblock MIT Press, Cambridge, MA.

\bibitem[{Chomsky(1986)}]{chomsky_knowledge_1986}
N.~Chomsky. 1986.
\newblock \emph{Knowledge of language: {Its} nature, origin, and use}.
\newblock Praeger Publishers.

\bibitem[{Chomsky et~al.(2023)Chomsky, Roberts, and Watumull}]{chomsky2023nyt}
Noam Chomsky, Ian Roberts, and Jeffrey Watumull. 2023.
\newblock \href {https://www.nytimes.com/2023/03/08/opinion/noam-chomsky-chatgpt-ai.html} {{Noam {Chomsky}: The False Promise of {ChatGPT}}}.
\newblock \emph{The New York Times}.

\bibitem[{Dalrymple and King(2019)}]{dalrymple_amazing_2019}
Mary Dalrymple and Tracy~Holloway King. 2019.
\newblock An amazing four doctoral dissertations.
\newblock \emph{Argumentum}, 15(2019).
\newblock Publisher: Debreceni Egyetemi Kiado.

\bibitem[{Eldan and Li(2023)}]{eldan2023tinystories}
Ronen Eldan and Yuanzhi Li. 2023.
\newblock {TinyStories: How Small Can Language Models Be and Still Speak Coherent English?}
\newblock \emph{arXiv:2305.07759}.

\bibitem[{Futrell et~al.(2019)Futrell, Wilcox, Morita, Qian, Ballesteros, and Levy}]{futrell-etal-2019-neural}
Richard Futrell, Ethan Wilcox, Takashi Morita, Peng Qian, Miguel Ballesteros, and Roger Levy. 2019.
\newblock \href {https://doi.org/10.18653/v1/N19-1004} {Neural language models as psycholinguistic subjects: Representations of syntactic state}.
\newblock In \emph{Proceedings of the 2019 Conference of the North {A}merican Chapter of the Association for Computational Linguistics: Human Language Technologies, Volume 1 (Long and Short Papers)}, pages 32--42, Minneapolis, Minnesota. Association for Computational Linguistics.

\bibitem[{Gerlach and Font-Clos(2020)}]{gerlach2020standardized}
Martin Gerlach and Francesc Font-Clos. 2020.
\newblock {A standardized Project Gutenberg corpus for statistical analysis of natural language and quantitative linguistics}.
\newblock \emph{Entropy}, 22(1):126.

\bibitem[{Goldberg(1995)}]{goldberg1995constructions}
Adele~E Goldberg. 1995.
\newblock \emph{Constructions: A Construction Grammar Approach to Argument Structure}.
\newblock University of Chicago Press.

\bibitem[{Goldberg(2005)}]{goldberg2005constructions}
Adele~E Goldberg. 2005.
\newblock \emph{Constructions at Work: The Nature of Generalization in Language}.
\newblock Oxford University Press.

\bibitem[{Goldberg(2019)}]{goldberg2019explain}
Adele~E Goldberg. 2019.
\newblock \emph{Explain me this: Creativity, competition, and the partial productivity of constructions}.
\newblock Princeton University Press.

\bibitem[{Heafield(2011)}]{heafield-2011-kenlm}
Kenneth Heafield. 2011.
\newblock \href {https://aclanthology.org/W11-2123} {{K}en{LM}: Faster and smaller language model queries}.
\newblock In \emph{Proceedings of the Sixth Workshop on Statistical Machine Translation}, pages 187--197, Edinburgh, Scotland. Association for Computational Linguistics.

\bibitem[{Hill et~al.(2016)Hill, Bordes, Chopra, and Weston}]{hill2015goldilocks}
Felix Hill, Antoine Bordes, Sumit Chopra, and Jason Weston. 2016.
\newblock \href {https://arxiv.org/abs/1511.02301} {{The Goldilocks Principle: Reading Children’s Books with Explicit Memory Representations}}.
\newblock In \emph{4th International Conference on Learning Representations, ICLR 2016}.

\bibitem[{Honnibal et~al.(2020)Honnibal, Montani, Van~Landeghem, and Boyd}]{spacy}
Matthew Honnibal, Ines Montani, Sofie Van~Landeghem, and Adriane Boyd. 2020.
\newblock \href {https://doi.org/10.5281/zenodo.1212303} {{spaCy}: Industrial-strength natural language processing in python}.

\bibitem[{Huebner et~al.(2021)Huebner, Sulem, Cynthia, and Roth}]{huebner-etal-2021-babyberta}
Philip~A. Huebner, Elior Sulem, Fisher Cynthia, and Dan Roth. 2021.
\newblock \href {https://doi.org/10.18653/v1/2021.conll-1.49} {{B}aby{BERT}a: Learning more grammar with small-scale child-directed language}.
\newblock In \emph{Proceedings of the 25th Conference on Computational Natural Language Learning}, pages 624--646, Online. Association for Computational Linguistics.

\bibitem[{Jumelet et~al.(2021)Jumelet, Denic, Szymanik, Hupkes, and Steinert-Threlkeld}]{jumelet-etal-2021-language}
Jaap Jumelet, Milica Denic, Jakub Szymanik, Dieuwke Hupkes, and Shane Steinert-Threlkeld. 2021.
\newblock \href {https://doi.org/10.18653/v1/2021.findings-acl.439} {Language models use monotonicity to assess {NPI} licensing}.
\newblock In \emph{Findings of the Association for Computational Linguistics: ACL-IJCNLP 2021}, pages 4958--4969, Online. Association for Computational Linguistics.

\bibitem[{Kallini et~al.(2024)Kallini, Papadimitriou, Futrell, Mahowald, and Potts}]{kallini2024mission}
Julie Kallini, Isabel Papadimitriou, Richard Futrell, Kyle Mahowald, and Christopher Potts. 2024.
\newblock Mission: Impossible language models.
\newblock \emph{arXiv:2401.06416}.

\bibitem[{Kayne(2007)}]{kayne2007syntax}
Richard~S Kayne. 2007.
\newblock {On the syntax of quantity in English}.
\newblock \emph{Linguistic theory and South Asian languages: Essays in honour of Ka Jayaseelan}, 102:73.

\bibitem[{Keenan(2013)}]{keenan_pleasant_2013}
Caitlin Keenan. 2013.
\newblock ``{A} pleasant three days in {Philadelphia}'': {Arguments} for a pseudopartitive analysis.
\newblock \emph{University of Pennsylvania Working Papers in Linguistics}, 19(1):11.

\bibitem[{Kim et~al.(2022)Kim, Linzen, and Smolensky}]{kim2022uncontrolled}
Najoung Kim, Tal Linzen, and Paul Smolensky. 2022.
\newblock \href {https://arxiv.org/abs/2212.10769} {{Uncontrolled Lexical Exposure Leads to Overestimation of Compositional Generalization in Pretrained Models}}.
\newblock \emph{arXiv:2212.10769}.

\bibitem[{Lau et~al.(2017)Lau, Clark, and Lappin}]{lau2017grammaticality}
Jey~Han Lau, Alexander Clark, and Shalom Lappin. 2017.
\newblock Grammaticality, acceptability, and probability: A probabilistic view of linguistic knowledge.
\newblock \emph{Cognitive Science}, 41(5):1202--1241.

\bibitem[{Leong and Linzen(2023)}]{leong-linzen-2023-language}
Cara Su-Yi Leong and Tal Linzen. 2023.
\newblock \href {https://aclanthology.org/2023.scil-1.11} {Language models can learn exceptions to syntactic rules}.
\newblock In \emph{Proceedings of the Society for Computation in Linguistics 2023}, pages 133--144, Amherst, MA. Association for Computational Linguistics.

\bibitem[{Leong and Linzen(2024)}]{leong2024testing}
Cara Su-Yi Leong and Tal Linzen. 2024.
\newblock Testing learning hypotheses using neural networks by manipulating learning data.
\newblock \emph{arXiv:2407.04593}.

\bibitem[{Li et~al.(2022)Li, Zhu, Thomas, Rudzicz, and Xu}]{li-etal-2022-neural}
Bai Li, Zining Zhu, Guillaume Thomas, Frank Rudzicz, and Yang Xu. 2022.
\newblock \href {https://doi.org/10.18653/v1/2022.acl-long.512} {Neural reality of argument structure constructions}.
\newblock In \emph{Proceedings of the 60th Annual Meeting of the Association for Computational Linguistics (Volume 1: Long Papers)}, pages 7410--7423, Dublin, Ireland. Association for Computational Linguistics.

\bibitem[{Linzen(2020)}]{linzen-2020-accelerate}
Tal Linzen. 2020.
\newblock \href {https://doi.org/10.18653/v1/2020.acl-main.465} {How can we accelerate progress towards human-like linguistic generalization?}
\newblock In \emph{Proceedings of the 58th Annual Meeting of the Association for Computational Linguistics}, pages 5210--5217, Online. Association for Computational Linguistics.

\bibitem[{Linzen et~al.(2016)Linzen, Dupoux, and Goldberg}]{linzen-etal-2016-assessing}
Tal Linzen, Emmanuel Dupoux, and Yoav Goldberg. 2016.
\newblock \href {https://doi.org/10.1162/tacl_a_00115} {Assessing the ability of {LSTM}s to learn syntax-sensitive dependencies}.
\newblock \emph{Transactions of the Association for Computational Linguistics}, 4:521--535.

\bibitem[{Lison and Tiedemann(2016)}]{lison-tiedemann-2016-opensubtitles2016}
Pierre Lison and J{\"o}rg Tiedemann. 2016.
\newblock \href {https://aclanthology.org/L16-1147} {{O}pen{S}ubtitles2016: Extracting large parallel corpora from movie and {TV} subtitles}.
\newblock In \emph{Proceedings of the Tenth International Conference on Language Resources and Evaluation ({LREC}'16)}, pages 923--929, Portoro{\v{z}}, Slovenia. European Language Resources Association (ELRA).

\bibitem[{Liu et~al.(2019)Liu, Ott, Goyal, Du, Joshi, Chen, Levy, Lewis, Zettlemoyer, and Stoyanov}]{liu2019roberta}
Yinhan Liu, Myle Ott, Naman Goyal, Jingfei Du, Mandar Joshi, Danqi Chen, Omer Levy, Mike Lewis, Luke Zettlemoyer, and Veselin Stoyanov. 2019.
\newblock \href {https://arxiv.org/abs/1907.11692} {{RoBERTa: A Robustly Optimized BERT Pretraining Approach}}.
\newblock \emph{arXiv:1907.11692}.

\bibitem[{Lu et~al.(2020)Lu, Mardziel, Wu, Amancharla, and Datta}]{lu2020gender}
Kaiji Lu, Piotr Mardziel, Fangjing Wu, Preetam Amancharla, and Anupam Datta. 2020.
\newblock \href {https://link.springer.com/chapter/10.1007/978-3-030-62077-6_14} {Gender bias in neural natural language processing}.
\newblock \emph{{Logic, Language, and Security: Essays Dedicated to Andre Scedrov on the Occasion of His 65th Birthday}}, pages 189--202.

\bibitem[{MacWhinney(2000)}]{macwhinney_childes_2000}
B.~MacWhinney. 2000.
\newblock \emph{The {CHILDES} project: {Tools} for analyzing talk}.
\newblock Lawrence Erlbaum Hillsdale, New Jersey.

\bibitem[{Mahowald(2023)}]{mahowald-2023-discerning}
Kyle Mahowald. 2023.
\newblock \href {https://doi.org/10.18653/v1/2023.eacl-main.20} {A discerning several thousand judgments: {GPT}-3 rates the article + adjective + numeral + noun construction}.
\newblock In \emph{Proceedings of the 17th Conference of the European Chapter of the Association for Computational Linguistics}, pages 265--273, Dubrovnik, Croatia. Association for Computational Linguistics.

\bibitem[{Mahowald et~al.(2024)Mahowald, Ivanova, Blank, Kanwisher, Tenenbaum, and Fedorenko}]{mahowald2024dissociating}
Kyle Mahowald, Anna~A Ivanova, Idan~A Blank, Nancy Kanwisher, Joshua~B Tenenbaum, and Evelina Fedorenko. 2024.
\newblock Dissociating language and thought in large language models.
\newblock \emph{Trends in Cognitive Sciences}.

\bibitem[{Maudslay et~al.(2019)Maudslay, Gonen, Cotterell, and Teufel}]{hall-maudslay-etal-2019-name}
Rowan~Hall Maudslay, Hila Gonen, Ryan Cotterell, and Simone Teufel. 2019.
\newblock \href {https://doi.org/10.18653/v1/D19-1530} {It{'}s all in the name: Mitigating gender bias with name-based counterfactual data substitution}.
\newblock In \emph{Proceedings of the 2019 Conference on Empirical Methods in Natural Language Processing and the 9th International Joint Conference on Natural Language Processing (EMNLP-IJCNLP)}, pages 5267--5275, Hong Kong, China. Association for Computational Linguistics.

\bibitem[{McCoy et~al.(2023)McCoy, Smolensky, Linzen, Gao, and Celikyilmaz}]{mccoy-etal-2023-much}
R.~Thomas McCoy, Paul Smolensky, Tal Linzen, Jianfeng Gao, and Asli Celikyilmaz. 2023.
\newblock \href {https://doi.org/10.1162/tacl_a_00567} {How much do language models copy from their training data? evaluating linguistic novelty in text generation using {RAVEN}}.
\newblock \emph{Transactions of the Association for Computational Linguistics}, 11:652--670.

\bibitem[{Misra(2022)}]{misra2022minicons}
Kanishka Misra. 2022.
\newblock \href {https://arxiv.org/abs/2203.13112} {minicons: Enabling flexible behavioral and representational analyses of transformer language models}.
\newblock \emph{arXiv:2203.13112}.

\bibitem[{Misra and Kim(2023)}]{misra-and-kim-2023-catabs}
Kanishka Misra and Najoung Kim. 2023.
\newblock \href {https://arxiv.org/abs/2312.03708} {{Abstraction via exemplars? A representational case study on lexical category inference in BERT}}.
\newblock In \emph{BUCLD 48: Proceedings of the 48th annual Boston University Conference on Language Development}, Boston, USA.

\bibitem[{O'Donnell(2015)}]{o2015productivity}
Timothy~J O'Donnell. 2015.
\newblock \emph{Productivity and reuse in language: A theory of linguistic computation and storage}.
\newblock MIT Press.

\bibitem[{Osherson et~al.(1990)Osherson, Smith, Wilkie, Lopez, and Shafir}]{osherson1990category}
Daniel~N Osherson, Edward~E Smith, Ormond Wilkie, Alejandro Lopez, and Eldar Shafir. 1990.
\newblock {Category-based Induction}.
\newblock \emph{Psychological Review}, 97(2):185.

\bibitem[{Patil et~al.(2024)Patil, Jumelet, Chiu, Lapastora, Shen, Wang, Willrich, and Steinert-Threlkeld}]{patil2024filtered}
Abhinav Patil, Jaap Jumelet, Yu~Ying Chiu, Andy Lapastora, Peter Shen, Lexie Wang, Clevis Willrich, and Shane Steinert-Threlkeld. 2024.
\newblock \href {https://arxiv.org/abs/2405.15750} {{Filtered Corpus Training (FiCT) Shows that Language Models can Generalize from Indirect Evidence}}.
\newblock \emph{arXiv:2405.15750}.

\bibitem[{Pauls and Klein(2012)}]{pauls-klein-2012-large}
Adam Pauls and Dan Klein. 2012.
\newblock \href {https://aclanthology.org/P12-1101} {Large-scale syntactic language modeling with treelets}.
\newblock In \emph{Proceedings of the 50th Annual Meeting of the Association for Computational Linguistics (Volume 1: Long Papers)}, pages 959--968, Jeju Island, Korea. Association for Computational Linguistics.

\bibitem[{Pearl(2022)}]{pearl2022poverty}
Lisa Pearl. 2022.
\newblock \href {https://sites.socsci.uci.edu/~lpearl/courses/readings/Pearl2019Ms_PovStimWithoutTears.pdf} {Poverty of the stimulus without tears}.
\newblock \emph{Language Learning and Development}, 18(4):415--454.

\bibitem[{Potts(2023)}]{potts2023pipps}
Christopher Potts. 2023.
\newblock \href {https://lingbuzz.net/lingbuzz/007495} {Characterizing {English} {Preposing} in {PP} constructions}.
\newblock Ms., Stanford University.

\bibitem[{Prange et~al.(2021)Prange, Schneider, and Srikumar}]{prange-etal-2021-supertagging}
Jakob Prange, Nathan Schneider, and Vivek Srikumar. 2021.
\newblock \href {https://doi.org/10.1162/tacl_a_00364} {Supertagging the long tail with tree-structured decoding of complex categories}.
\newblock \emph{Transactions of the Association for Computational Linguistics}, 9:243--260.

\bibitem[{Pullum(2017)}]{pullum2017theory}
Geoffrey~K Pullum. 2017.
\newblock Theory, data, and the epistemology of syntax.
\newblock In \emph{Grammatische Variation. Empirische Zug{\"a}nge und theoretische Modellierung}, pages 283--298. de Gruyter.

\bibitem[{Pullum and Scholz(2002)}]{pullum2002empirical}
Geoffrey~K Pullum and Barbara~C Scholz. 2002.
\newblock Empirical assessment of stimulus poverty arguments.
\newblock \emph{The {L}inguistic {R}eview}, 19(1-2):9--50.

\bibitem[{Radford et~al.(2019)Radford, Wu, Child, Luan, Amodei, and Sutskever}]{radford2019language}
Alec Radford, Jeff Wu, Rewon Child, David Luan, Dario Amodei, and Ilya Sutskever. 2019.
\newblock \href {https://d4mucfpksywv.cloudfront.net/better-language-models/language_models_are_unsupervised_multitask_learners.pdf} {Language models are unsupervised multitask learners}.
\newblock \emph{OpenAI}.

\bibitem[{Schwarzschild(2011)}]{schwarzschild_stubborn_2011}
Roger Schwarzschild. 2011.
\newblock \href {https://semantics.uchicago.edu/kennedy/classes/f11/na/docs/schwarzschild09.pdf} {{Stubborn Distributivity, Multiparticipant Nouns and the Count/Mass Distinction}}.
\newblock In \emph{Proceedings of {NELS}}, volume~39, pages 661--678. Graduate Linguistics Students Association, University of Massachusetts.
\newblock Issue: 2.

\bibitem[{Sinha et~al.(2021)Sinha, Jia, Hupkes, Pineau, Williams, and Kiela}]{sinha-etal-2021-masked}
Koustuv Sinha, Robin Jia, Dieuwke Hupkes, Joelle Pineau, Adina Williams, and Douwe Kiela. 2021.
\newblock \href {https://doi.org/10.18653/v1/2021.emnlp-main.230} {Masked language modeling and the distributional hypothesis: Order word matters pre-training for little}.
\newblock In \emph{Proceedings of the 2021 Conference on Empirical Methods in Natural Language Processing}, pages 2888--2913, Online and Punta Cana, Dominican Republic. Association for Computational Linguistics.

\bibitem[{Solt(2007)}]{solt_two_2007}
Stephanie Solt. 2007.
\newblock Two types of modified cardinals.
\newblock In \emph{International {Conference} on {Adjectives}. {Lille}}.

\bibitem[{Stolcke et~al.(2000)Stolcke, Ries, Coccaro, Shriberg, Bates, Jurafsky, Taylor, Martin, Van Ess-Dykema, and Meteer}]{stolcke-etal-2000-dialogue}
Andreas Stolcke, Klaus Ries, Noah Coccaro, Elizabeth Shriberg, Rebecca Bates, Daniel Jurafsky, Paul Taylor, Rachel Martin, Carol Van Ess-Dykema, and Marie Meteer. 2000.
\newblock \href {https://aclanthology.org/J00-3003} {Dialogue act modeling for automatic tagging and recognition of conversational speech}.
\newblock \emph{Computational Linguistics}, 26(3):339--374.

\bibitem[{Suttle and Goldberg(2011)}]{suttle2011partial}
Laura Suttle and Adele~E Goldberg. 2011.
\newblock \href {https://adele.princeton.edu/wp-content/uploads/sites/277/2015/01/SuttleGoldbergLinguistics11.pdf} {The partial productivity of constructions as induction}.
\newblock \emph{Linguistics}, 49(6):1237--1269.

\bibitem[{Tayyar~Madabushi et~al.(2020)Tayyar~Madabushi, Romain, Divjak, and Milin}]{tayyar-madabushi-etal-2020-cxgbert}
Harish Tayyar~Madabushi, Laurence Romain, Dagmar Divjak, and Petar Milin. 2020.
\newblock \href {https://doi.org/10.18653/v1/2020.coling-main.355} {{C}x{GBERT}: {BERT} meets construction grammar}.
\newblock In \emph{Proceedings of the 28th International Conference on Computational Linguistics}, pages 4020--4032, Barcelona, Spain (Online). International Committee on Computational Linguistics.

\bibitem[{Touvron et~al.(2023)Touvron, Martin, Stone, Albert, Almahairi, Babaei, Bashlykov, Batra, Bhargava, Bhosale et~al.}]{touvron2023llama}
Hugo Touvron, Louis Martin, Kevin Stone, Peter Albert, Amjad Almahairi, Yasmine Babaei, Nikolay Bashlykov, Soumya Batra, Prajjwal Bhargava, Shruti Bhosale, et~al. 2023.
\newblock \href {https://arxiv.org/abs/2307.09288} {{Llama 2: Open foundation and fine-tuned chat models}}.
\newblock \emph{arXiv:2307.09288}.

\bibitem[{Tseng et~al.(2022)Tseng, Shih, Chen, Chou, Ku, and Hsieh}]{tseng-etal-2022-cxlm}
Yu-Hsiang Tseng, Cing-Fang Shih, Pin-Er Chen, Hsin-Yu Chou, Mao-Chang Ku, and Shu-Kai Hsieh. 2022.
\newblock \href {https://aclanthology.org/2022.lrec-1.683} {{C}x{LM}: A construction and context-aware language model}.
\newblock In \emph{Proceedings of the Thirteenth Language Resources and Evaluation Conference}, pages 6361--6369, Marseille, France. European Language Resources Association.

\bibitem[{Veenboer and Bloem(2023)}]{veenboer-bloem-2023-using}
Tim Veenboer and Jelke Bloem. 2023.
\newblock \href {https://doi.org/10.18653/v1/2023.findings-acl.819} {Using collostructional analysis to evaluate {BERT}{'}s representation of linguistic constructions}.
\newblock In \emph{Findings of the Association for Computational Linguistics: ACL 2023}, pages 12937--12951, Toronto, Canada. Association for Computational Linguistics.

\bibitem[{Warstadt(2022)}]{warstadt2022artificial}
Alex Warstadt. 2022.
\newblock \emph{Artificial Neural Networks as Models of Human Language Acquisition}.
\newblock New York University.

\bibitem[{Warstadt et~al.(2023)Warstadt, Mueller, Choshen, Wilcox, Zhuang, Ciro, Mosquera, Paranjabe, Williams, Linzen, and Cotterell}]{warstadt-etal-2023-findings}
Alex Warstadt, Aaron Mueller, Leshem Choshen, Ethan Wilcox, Chengxu Zhuang, Juan Ciro, Rafael Mosquera, Bhargavi Paranjabe, Adina Williams, Tal Linzen, and Ryan Cotterell. 2023.
\newblock \href {https://doi.org/10.18653/v1/2023.conll-babylm.1} {Findings of the {B}aby{LM} challenge: Sample-efficient pretraining on developmentally plausible corpora}.
\newblock In \emph{Proceedings of the BabyLM Challenge at the 27th Conference on Computational Natural Language Learning}, pages 1--34, Singapore. Association for Computational Linguistics.

\bibitem[{Weber et~al.(2021)Weber, Jumelet, Bruni, and Hupkes}]{weber-etal-2021-language}
Lucas Weber, Jaap Jumelet, Elia Bruni, and Dieuwke Hupkes. 2021.
\newblock \href {https://doi.org/10.18653/v1/2021.eacl-main.176} {Language modelling as a multi-task problem}.
\newblock In \emph{Proceedings of the 16th Conference of the European Chapter of the Association for Computational Linguistics: Main Volume}, pages 2049--2060, Online. Association for Computational Linguistics.

\bibitem[{Wei et~al.(2021)Wei, Garrette, Linzen, and Pavlick}]{wei-etal-2021-frequency}
Jason Wei, Dan Garrette, Tal Linzen, and Ellie Pavlick. 2021.
\newblock \href {https://doi.org/10.18653/v1/2021.emnlp-main.72} {Frequency effects on syntactic rule learning in transformers}.
\newblock In \emph{Proceedings of the 2021 Conference on Empirical Methods in Natural Language Processing}, pages 932--948, Online and Punta Cana, Dominican Republic. Association for Computational Linguistics.

\bibitem[{Weissweiler et~al.(2022)Weissweiler, Hofmann, K{\"o}ksal, and Sch{\"u}tze}]{weissweiler-etal-2022-better}
Leonie Weissweiler, Valentin Hofmann, Abdullatif K{\"o}ksal, and Hinrich Sch{\"u}tze. 2022.
\newblock \href {https://doi.org/10.18653/v1/2022.emnlp-main.746} {The better your syntax, the better your semantics? probing pretrained language models for the {E}nglish comparative correlative}.
\newblock In \emph{Proceedings of the 2022 Conference on Empirical Methods in Natural Language Processing}, pages 10859--10882, Abu Dhabi, United Arab Emirates. Association for Computational Linguistics.

\bibitem[{Weissweiler et~al.(2024)Weissweiler, K{\"o}ksal, and Sch{\"u}tze}]{weissweiler2024hybrid}
Leonie Weissweiler, Abdullatif K{\"o}ksal, and Hinrich Sch{\"u}tze. 2024.
\newblock Hybrid human-{LLM} corpus construction and {LLM} evaluation for rare linguistic phenomena.
\newblock \emph{arXiv:2403.06965}.

\bibitem[{Wilcox et~al.(2018)Wilcox, Levy, Morita, and Futrell}]{wilcox-etal-2018-rnn}
Ethan Wilcox, Roger Levy, Takashi Morita, and Richard Futrell. 2018.
\newblock \href {https://doi.org/10.18653/v1/W18-5423} {What do {RNN} language models learn about filler{--}gap dependencies?}
\newblock In \emph{Proceedings of the 2018 {EMNLP} Workshop {B}lackbox{NLP}: Analyzing and Interpreting Neural Networks for {NLP}}, pages 211--221, Brussels, Belgium. Association for Computational Linguistics.

\bibitem[{Wolf et~al.(2020)Wolf, Debut, Sanh, Chaumond, Delangue, Moi, Cistac, Rault, Louf, Funtowicz, Davison, Shleifer, von Platen, Ma, Jernite, Plu, Xu, Le~Scao, Gugger, Drame, Lhoest, and Rush}]{wolf-etal-2020-transformers}
Thomas Wolf, Lysandre Debut, Victor Sanh, Julien Chaumond, Clement Delangue, Anthony Moi, Pierric Cistac, Tim Rault, Remi Louf, Morgan Funtowicz, Joe Davison, Sam Shleifer, Patrick von Platen, Clara Ma, Yacine Jernite, Julien Plu, Canwen Xu, Teven Le~Scao, Sylvain Gugger, Mariama Drame, Quentin Lhoest, and Alexander Rush. 2020.
\newblock \href {https://doi.org/10.18653/v1/2020.emnlp-demos.6} {Transformers: State-of-the-art natural language processing}.
\newblock In \emph{Proceedings of the 2020 Conference on Empirical Methods in Natural Language Processing: System Demonstrations}, pages 38--45, Online. Association for Computational Linguistics.

\bibitem[{Xu and Tenenbaum(2007)}]{xu2007word}
Fei Xu and Joshua~B Tenenbaum. 2007.
\newblock {Word learning as Bayesian inference}.
\newblock \emph{Psychological Review}, 114(2):245.

\bibitem[{Zhang et~al.(2022)Zhang, Roller, Goyal, Artetxe, Chen, Chen, Dewan, Diab, Li, Lin et~al.}]{zhang2022opt}
Susan Zhang, Stephen Roller, Naman Goyal, Mikel Artetxe, Moya Chen, Shuohui Chen, Christopher Dewan, Mona Diab, Xian Li, Xi~Victoria Lin, et~al. 2022.
\newblock \href {https://arxiv.org/abs/2205.01068} {{OPT: Open Pre-trained Transformer Language Models}}.
\newblock \emph{arXiv:2205.01068}.

\end{thebibliography}

\appendix

\section{Dataset Access and Licensing}
The \aann{} acceptability dataset by \citet{mahowald-2023-discerning} is released using the MIT License and  was accessed from the author's public github repo.\footnote{\url{https://github.com/mahowak/aann-public}}
The BabyLM dataset\footnote{accessed from \url{https://babylm.github.io/}} does not have a single license of its own but instead inherits the licenses of its constituents: CHILDES \citep{macwhinney_childes_2000}, BNC Dialogue portion,\footnote{\url{http://www.natcorp.ox.ac.uk}} Children's Book Test \citep{hill2015goldilocks}, Children's Stories Text Corpus,\footnote{\url{https://www.kaggle.com/datasets/edenbd/children-stories-text-corpus}} Standardized Project Gutenberg Corpus \citep{gerlach2020standardized}, OpenSubtitles \citep{lison-tiedemann-2016-opensubtitles2016}, QCRI Educational Domain Corpus \citep{abdelali-etal-2014-amara}, Wikipedia,\footnote{\url{https://dumps.wikimedia.org/enwiki/20221220/}} Simple Wikipedia,\footnote{\url{https://dumps.wikimedia.org/simplewiki/20221201/}} Switchboard Dialog Act Corpus \citep{stolcke-etal-2000-dialogue}. 
Since this dataset was specifically released to train LMs, we work under the assumption that our LMs do not violate its license policies. 
We will follow the inherited licenses' policies while making the trained LMs and ablated BabyLM data public, and refrain from releasing them if we find them to be in violation of the policies.

\section{LM training details}
\label{sec:trainingdetails}
As mentioned in the main text (see \Cref{sec:method}), we use the OPT architecture \citep{zhang2022opt} to train our LMs on all versions of the BabyLM corpus. This was the best performing autoregressive LM in the BabyLM Competition \citep{warstadt-etal-2023-findings}. 
For each instance of the BabyLM (ablated or otherwise), we tune the learning rate\footnote{We searched the following set: \texttt{\{1e-4, 3e-4, 1e-3, 3e-3\}}} based on the validation set, and use the best learning rate as a result of the tuning to train an additional two language models using different seeds. As a result, for each ablation of the BabyLM corpus, we run 6 LM training experiments, which amounts to a whopping 114 LMs for all our experiments.
\Cref{tab:trainingdetails} contains further details of the training. 

\begin{table}[h]
\centering
\begin{tabular}{@{}lr@{}}
\toprule
\textbf{(Hyper)parameter} & \textbf{Value} \\ \midrule
Architecture & OPT \citep{zhang2022opt} \\
Embed size & 768 \\
FFN dimension & 3,072 \\
Num. layers & 12 \\
Attention heads & 12 \\
Vocab size & 16,384 \\
Max. seq. length & 256 \\
Batch size & 32 \\
Warmup steps & 32,000 \\
Epochs & 20 \\
Total parameters & 97M \\
Training size & 100M tokens \\
Compute & 1x NVIDIA A40 \\
Training time & 21 hours \\ \bottomrule
\end{tabular}%
\caption{LM Training details}
\label{tab:trainingdetails}
\end{table}

\section{Detecting \aanns{} and related phenomena}
\label{sec:detection}

In this section, we briefly describe our methods to extract constructions and phenomena relevant to this paper from the BabyLM corpus \citep{warstadt-etal-2023-findings}.
Our methods primarily rely on: 1) the surface form of the sentences in the corpus; 2) their corresponding part-of-speech (POS) tag sequences; and in a few cases, 3) their dependency parses. For the latter two, we used \texttt{spacy} \citep{spacy}, specifically, its \texttt{en\_web\_trf} model, which is based on the RoBERTa-base LM \citep{liu2019roberta}.
Next we describe how we used these artifacts to detect our target constructions:

\subsection{\aanns}

To detect \aanns{}, we constructed a regex-based pattern-matcher which operated over a POS-tagged version of the BabyLM corpus. 
We started with an initial regex pattern (\textcolor{regexv1}{\texttt{Regex v1}}), as shown in \Cref{regexv1}:

\begin{figure}
    \centering
    \includegraphics[width=\columnwidth]{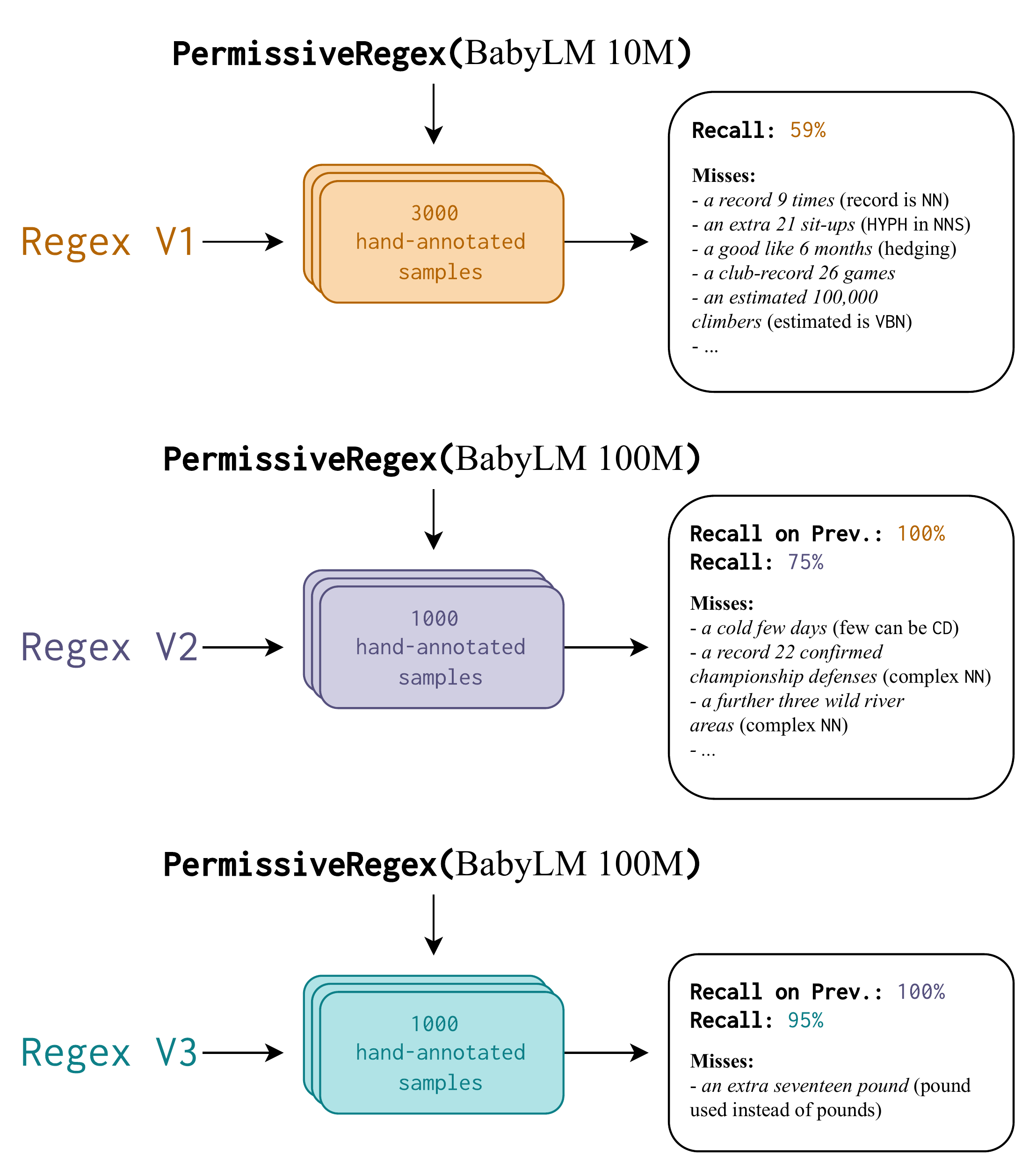}
    \caption{Pipeline to assess the recall of our \aann{}-detecting regex patterns, along with examples of cases missed by each regex. The recall for our final regex (\textcolor{regexv3}{\texttt{Regex v3}}) is 95\% (missing only one instance where there was a typo), and it is able to handle complex and sophisticated forms of the construction.}
    \label{fig:recall-pipeline}
\end{figure}

\begin{lstlisting}[language=Python,stringstyle=\ttfamily\color{regexv1},caption={Regex v1.},label={regexv1}]
pattern = r'\b(DT)(?:(?:\s(RB))*\s(JJ|JJR|JJS)(?:\s(CC))*)+(\s(CD|JJ|JJR|JJS|NN|CD\sCD)(?:\s(TO|CC)\s(CD))*)(\s(NNS|NNPS|(NN\sNNS)|((NN|NNS) IN NNS)))+'
\end{lstlisting}

\noindent
here we restrict the determiner (\texttt{DT}) to be either `a', `an', or `another'.
This regex permits multiple adjectives (\textit{an \textbf{exhilarating and marvelous} three months}) optional adverbs (\textit{an \textbf{excruciatingly} painful two semesters}), multi-word noun phrases with plural head-nouns (\textit{a refreshing two \textbf{glasses of aperol spritz}}), numeral-expressions involving subordinate clauses (\textit{a measly \textbf{three to five} days}), among other potential edge cases.

We then tested this regex pattern on a large sample of utterances which we extracted using a permissive regex applied to the 10M-token version of BabyLM (a subset of our 100M training set), which looked for any ``a'' or ``an'' or ``another'' that appeared sequentially prior to a numeral as well as a plural noun in a sentence. Importantly this regex filter did not rely on any POS tagging, to avoid issues attributable to tagging errors. We hand-annotated a sample of 3000 utterances from this set, and found 49 legitimate \aanns{}.\footnote{In reality, we found 50, but rejected one of them: ``\textit{a good 1-2" of snow...}, where `"' is inches. This would have never been caught unless we are to include `"' in our pipeline which would conflate other uses of quotes.} Our \textcolor{regexv1}{\texttt{Regex v1}} only detected 29 of these, meaning its recall was around 59\%. 

We then developed a second version of the regex (\textcolor{prpl}{\texttt{Regex v2}}; see \cref{regexv2}) to handle cases that the above regex pattern missed (e.g., using participle modifiers, occurrence of punctuation or extra spaces in between, accounting for hedging, a case where `record' was used as a modifier, etc.). 

\begin{lstlisting}[language=Python,stringstyle=\ttfamily\color{prpl},caption={Regex v2.},label={regexv2}]
pattern = r'\bDT\s(((HYPH|,)\s))?((((RB|CC|IN)\s)+)?((JJ|JJR|JJS|VBN|((NN CC NN |NN HYPH )+(JJ|JJR|JJS|VBN)))((\s(HYPH|,))?)\s))+(((RB)\s)+)?(((HYPH|,)\s))?((UH)\s)?(((NN|CC)\s)+)?((CD)(\s(TO|CC|(HYPH|,))(\s(HYPH|,))?)?\s)+(((HYPH|,)\s))?(JJR\s)?(DT\s)?((NNS|NNPS|(NN\sNNS)|((NN|NNS) IN NNS)))+'
\end{lstlisting}

\noindent
To test \textcolor{prpl}{\texttt{Regex v2}}, we again used the permissive regex and extracted an additional 1000 samples from our training set. On hand-annotating them, we found 24 valid \aanns{}, out of which \textcolor{prpl}{\texttt{Regex v2}} detected 18, bringing up the recall to 75\%. 

In both the previous cases, we were post-processing the detected \aanns{} to include certain adjectives (\textit{few, dozen, couple, several, many, more}) as numerals, as per the guidelines of \citet{kayne2007syntax} and \citet{solt_two_2007}. This allows the following to also be considered instances of the \aann{}:

\ex. \label{ex:permittedaanns}
\a. a beautiful \textbf{few} days.
\b. an amazing \textbf{dozen} eggs.
\c. a pictorial \textbf{several} pages.
\d. a great \textbf{many} days.

At the same time, this also ends up including cases such as:

\ex. \label{ex:badaanns}
\a. a few hundred dollars. (few modifies hundred but not dollars)
\b. an awful couple of days. (pseudo-partitive)

Similarly, we had to include \texttt{NN} within our adjective span of the regex pattern to accommodate `record' when used as/as part of a modifier (e.g., \textit{a \textbf{record-high} 60 miles per hour}), but this exploded the number of ``detected'' \aanns{}, lowering our precision drastically, due to which we omitted it. 

To address these issues, we decided to pre-process the POS-tagged corpora prior to using our regex, where we substituted articles of interest with the `\texttt{ARTICLE}' tag, substituted \textit{record} when preceeded by an article with the `\texttt{RECORD}' tag, and numeral proxies with the `\texttt{FEW}' tag, though ensuring that it appeared linearly after a known adjective which was not a numeral proxy. This led to the creation of \textcolor{regexv3}{\texttt{Regex v3}} (\cref{finalregex}):

\begin{lstlisting}[language=Python,stringstyle=\ttfamily\color{regexv3},caption={Regex v3 (final). Tags such as \texttt{ARTICLE, RECORD, FEW} are added after POS-tagging to include certain special tokens.},label={finalregex}]
pattern = r'\bARTICLE\s(((HYPH|,)\s))?((((RB|CC|IN)\s)+)?((JJ|JJR|JJS|VBN|RECORD|((NN CC NN |NN HYPH )+(JJ|JJR|JJS|VBN|RECORD)))((\s(HYPH|,))?)\s))+(((RB)\s)+)?(((HYPH|,)\s))?((UH)\s)?(((NN|CC)\s)+)?((CD|FEW)(\s(TO|CC|(HYPH|,))(\s(HYPH|,))?)?\s)+(((HYPH|,)\s))?((JJR|JJ|VBN)\s)?(ARTICLE\s)?((NNS|NNPS|(NN\sNNS)|((NN|NNS) IN NNS)))+'
\end{lstlisting}

\noindent
This was able to handle the idiosyncracies of all previously detected \aanns{}. We again extracted a further additional 1000 samples to hand-annotate and found 18 attested \aanns{}. \textcolor{regexv3}{\texttt{Regex v3}} was able to detect 17 out of these (recall of 95\%), missing out on only one where an incorrect form was used in lieu of a plural noun (e.g., \textit{pound} instead of \textit{pounds}). We don't really consider this a meaningful missed example since the singular noun actually makes this a degenerate \aann, not a genuine one (but, to be conservative, count it as a miss for assessing a worst-case recall estimate).
At this point, we stopped further refining our regex and \textbf{used \textcolor{regexv3}{\texttt{Regex V3}} as our final detector}, while also acknowledging that it is perhaps impossible to guarantee whether every single \aann{} instance is captured by the regex. \Cref{fig:recall-pipeline} shows our recall analysis pipeline in a nutshell.

Once detected, we map the found constructions to their respective positions within the \aann{} format, which allows us to measure metrics such as slot variability, etc.

\subsection{DT ANNs}

We follow the exact same procedure as the one for \aanns{}, but no longer restrict the determiner position to only be an indefinite determiner.

\subsection{A few/couple/dozen NOUNs}
An important phenomenon that we consider to be related to the \aann{} involves cases such as: ``\textit{that only lasted a few days}'' and ``\textit{could you bring me a couple liters?}'', etc., where the plural nouns are attached to an indefinite article. To detect such cases, we consider the following two dependency configurations, where we have an indefinite determiner (\textit{a, an, another}) with either a \textcolor{OliveGreen}{\texttt{det}} relation with the plural noun (\texttt{NNS} or \texttt{NNPS}) or a \textcolor{RoyalBlue}{\texttt{quantmod}} relation with a noun which has a \textcolor{Orchid}{\texttt{nummod}} with the plural noun. In the former case, we usually have an \textcolor{BurntOrange}{\texttt{amod}} relation between the noun and the adjective.
\begin{center}
    \begin{dependency}
\begin{deptext}[column sep=0.6cm, row sep=.1ex]
\dots \& \texttt{DT} \& \texttt{JJ} \& \texttt{NNS} \& \dots \\
\dots \& a \& few \& days \& \dots \\
\end{deptext}
\depedge{2}{4}{\textcolor{OliveGreen}{det}}
\depedge{3}{4}{\textcolor{BurntOrange}{amod}}
\end{dependency}
\end{center}

\begin{center}
    \begin{dependency}
\begin{deptext}[column sep=0.6cm, row sep=.1ex]
\dots \& \texttt{DT} \& \texttt{NN} \& \texttt{NNS} \& \dots \\
\dots \& a \& couple \& days \& \dots \\
\end{deptext}
\depedge{2}{3}{\textcolor{RoyalBlue}{quantmod}}
\depedge{3}{4}{\textcolor{Orchid}{nummod}}
\end{dependency}
\end{center}

\subsection{Measure NNS with Singular Verbs}

Similar to the previous case, another phenomenon which might be related to the \aann{} constructions is when measure noun-phrases with plural nouns are treated as singular via their agreement with a verb---e.g., ``\textit{five dollars \textbf{is} plenty!}''
To detect such cases, we again rely on the following dependency configuration, where we have a plural noun (\texttt{NNS} or \texttt{NNPS}) attached to a cardinal number (\texttt{CD}) via the \textcolor{Orchid}{\texttt{nummod}} dependency relation, and at the same time also attached to singular verbs via the \textcolor{BrickRed}{\texttt{nsubj}} dependency relation (i.e., are subjects of singular verbs).

\begin{center}
    \begin{dependency}
\begin{deptext}[column sep=0.6cm]
\dots \& \texttt{CD} \& \texttt{NNS} \& \texttt{VB} \& \dots \\
\dots \& five \& dollars \& is \& \dots \\
\end{deptext}
\depedge{2}{3}{\textcolor{Orchid}{nummod}}
\depedge{3}{4}{\textcolor{BrickRed}{nsubj}}
\end{dependency}
\end{center}

\section{A/An + ADJ/NUM frequency balancing}

A corpus analysis of BabyLM, along with its POS-tagged version suggests that the sequence ``\texttt{a/an/another (JJ|JJR|JJS)}'' occurs 613,985 times while ``\texttt{a/an/another CD}'' occurs only 42,111 times -- this suggests that adjectives are approximately 14.6 more likely to follow an indefinite article than are numerals. We therefore balance these values by removing 571,874 instances where adjectives follow an indefinite article. This constitutes the largest-sized ablation in this work.

\section{Lexical semantic constraints on \aann{} slots}
\label{sec:lexicalsemantics}

\begin{figure}
    \centering
    \includegraphics[width=\columnwidth]{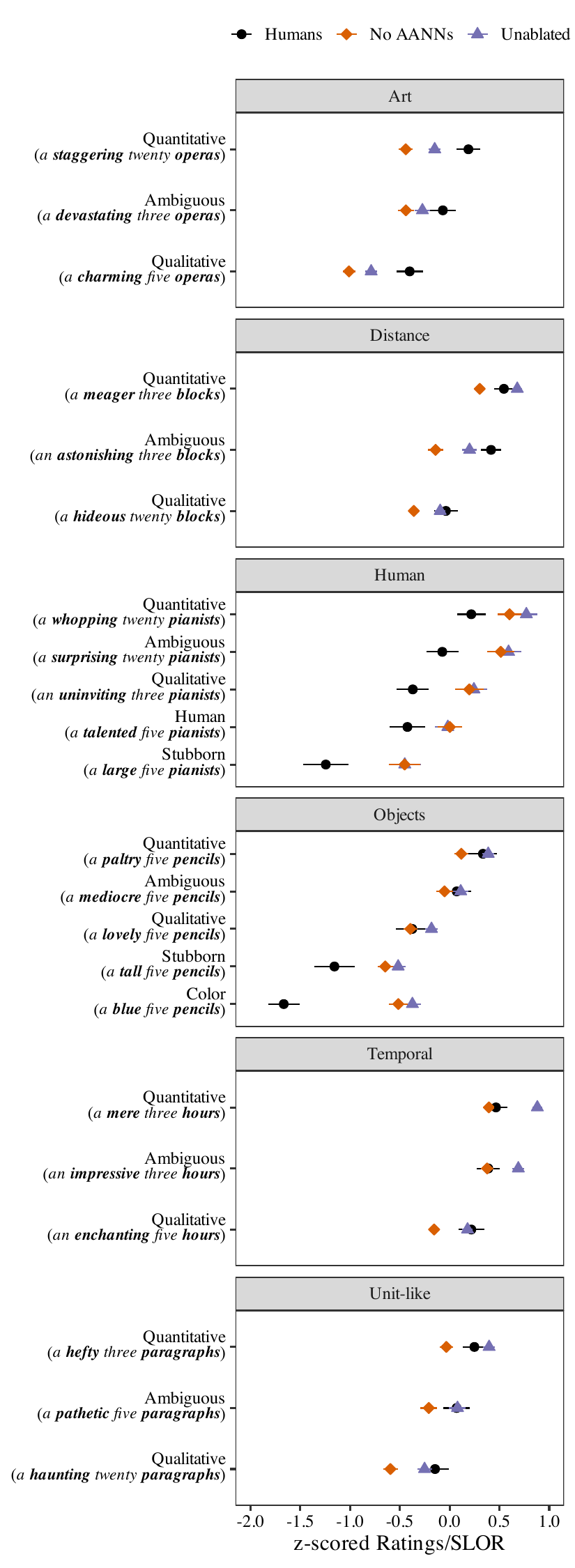}
    \caption{z-scored \aann{} acceptability ratings elicited from Humans (scale of 1-10) and LMs (\slor{}s) trained on corpora with (1) \textcolor{orang}{\aanns{} removed (i.e., \noaann{})}; and (2) \textcolor{prpl}{left unablated}. Ratings broken down based on adjective and noun classes. Ratings are computed for each system based on \citet{mahowald-2023-discerning}'s entire dataset, which consists of human derived acceptability judgments on 3,420 different types of \aanns{}.}
    \label{fig:lmhumans}
\end{figure}

\Cref{fig:lmhumans} shows the breakdown of acceptability ratings from humans and LMs across various adjective and noun classes.

\section{Variability Analysis}
\label{sec:va}

In \cref{sec:variability} we compared \aann{}-generalization of LMs trained on BabyLM versions which differed in the amount of variability that was present in the \aanns{} that the models were exposed to. In particular, we operationalized variability in terms of the slot-fillers of the adjective/numeral/noun slots, both together as well as individually. \Cref{tab:variability-examples} shows three examples of high and low variability items (each) for the four different slot-filler based considerations in our experiments.

\begin{table}[h]
\centering
\resizebox{\columnwidth}{!}{%
\begin{tabular}{@{}clrlr@{}}
\toprule
\multirow{2}{*}{\textbf{Slot}} & \multicolumn{2}{c}{\textbf{High Variability}} & \multicolumn{2}{c}{\textbf{Low Variability}} \\ \cmidrule(l){2-5} 
 & Instance & Freq. & Instance & Freq. \\ \midrule
\multirow{3}{*}{All} & impressive 30 appearances & 1 & great many things & 42 \\
 & massive 108 years & 1 & good many years & 21 \\
 & reported 14 million dolars & 1 & additional two years & 4 \\ \midrule
\multirow{3}{*}{Adj} & career-high & 38 & great & 355 \\
 & staggering & 12 & additional & 111 \\
 & measly & 7 & mere & 60 \\ \midrule
\multirow{3}{*}{Num} & 20 & 32 & two & 174 \\
 & couple & 17 & five & 67 \\
 & seven to eight & 1 & few & 64 \\ \midrule
\multirow{3}{*}{Noun} & dollars & 15 & years & 254 \\
 & students & 8 & miles & 77 \\
 & kangaroos & 1 & hours & 42 \\ \bottomrule
\end{tabular}%
}
\caption{Examples of slot fillers that were ablated as part of our variability experiments, along with their frequency in the training data, across all slots considered (All open slots, Adjective-only, Numeral-only, and Noun-only).}
\label{tab:variability-examples}
\end{table}

\end{document}